\documentclass[journal,twoside,web]{ieeecolor}
\usepackage{generic}
\usepackage{cite}
\usepackage{amsmath,amssymb,amsfonts}
\usepackage{algorithmic}
\usepackage{graphicx}
\usepackage{algorithm2e}
\usepackage{threeparttable}

\usepackage{subcaption}

\usepackage{wasysym}
\usepackage{booktabs}
\usepackage{makecell}
\usepackage{color}
\usepackage{wrapfig}
\usepackage{makecell}
\usepackage{array}
\usepackage{colortbl}
\definecolor{myblue}{RGB}{194,228,248}

\usepackage[misc]{ifsym}
\let\labelindent\relax
\usepackage{enumitem}
\setlist{nolistsep}

\usepackage{xcolor}

\usepackage{multirow}
\usepackage{colortbl}
\definecolor{mygray}{RGB}{240,240,240}

\usepackage{soul}
\definecolor{lightRed}{RGB}{250,189,194} 
\definecolor{DarkRed}{RGB}{199, 71, 68}
\definecolor{lightBlue}{RGB}{218, 227, 243}
\definecolor{DarkBlue}{RGB}{47, 84, 150}

\newcommand{\cRed}[1]{
  \begingroup
  \sethlcolor{lightRed}
  \textcolor{DarkRed}{\hl{#1}}
  \endgroup
}

\newcommand{\cBlue}[1]{
  \begingroup
  \sethlcolor{lightBlue}
  \textcolor{DarkBlue}{\hl{#1}}
  \endgroup
}

\usepackage{hyperref}
\hypersetup{hidelinks=true}
\usepackage{textcomp}
\def\BibTeX{{\rm B\kern-.05em{\sc i\kern-.025em b}\kern-.08em
    T\kern-.1667em\lower.7ex\hbox{E}\kern-.125emX}}
\begin{document}

\title{Enhancing 3D Medical Image Understanding with Pretraining Aided by 2D Multimodal Large Language Models}

\author{Qiuhui Chen, Xuancheng Yao, Huping Ye, and Yi Hong
\thanks{This research work was supported by the National Natural Science Foundation of China (NSFC) 62203303 and Shanghai Municipal Science and Technology Major Project 2021SHZDZX0102.
Qiuhui Chen, Xuancheng Yao, Huping Ye, and Yi Hong are with the School of Computer Science, Shanghai Jiao Tong University, Shanghai 200240, China (Corresponding author: Yi Hong, e-mail: yi.hong@sjtu.edu.cn).}
}

\maketitle

\begin{abstract}
Understanding 3D medical image volumes is critical in the medical field, yet existing 3D medical convolution and transformer-based self-supervised learning (SSL) methods often lack deep semantic comprehension. Recent advancements in multimodal large language models (MLLMs) provide a promising approach to enhance image understanding through text descriptions. To leverage these 2D MLLMs for improved 3D medical image understanding, we propose Med3DInsight, a novel pretraining framework that integrates 3D image encoders with 2D MLLMs via a specially designed plane-slice-aware transformer module. Additionally, our model employs a partial optimal transport based alignment, demonstrating greater tolerance to noise introduced by potential noises in LLM-generated content. Med3DInsight introduces a new paradigm for scalable multimodal 3D medical representation learning without requiring human annotations. Extensive experiments demonstrate our state-of-the-art performance on two downstream tasks, i.e., segmentation and classification, across various public datasets with CT and MRI modalities, outperforming current SSL methods. Med3DInsight can be seamlessly integrated into existing 3D medical image understanding networks, potentially enhancing their performance. Our source code, generated datasets, and pre-trained models will be available at \href{https://github.com/Qybc/Med3DInsight}{https://github.com/Qybc/Med3DInsight}.
\end{abstract}


\section{Introduction}
\label{sec:introduction}

\IEEEPARstart{I}{n} medical research, the ability to understand three-dimensional (3D) medical images is crucial for extracting critical information that informs diagnosis, treatment planning, and a variety of healthcare studies. With the advent of deep learning, researchers have developed specialized models for 3D image understanding tasks such as classification~\cite{chen2019med3d,jang2022m3t} 
and segmentation~\cite{butoi2023universeg,hatamizadeh2022unetr,zhou2023nnformer}.
While these task-specific models can perform well in their respective areas, they often lack versatility and require significant effort to design, train, and fine-tune for each task. In contrast, pretraining a unified model to learn comprehensive representations of 3D medical images offers the advantage of greater flexibility and efficiency. In particular, a pre-trained image understanding model can be fine-tuned for various downstream tasks, minimizing the necessity to develop separate models for each specific task. This approach leverages shared knowledge across diverse datasets, which collectively form a comprehensive large-scale set for pretraining the model. 

Pretraining methods like self-supervised learning that rely solely on images have revolutionized model training by eliminating the need for labeled data. Representative approaches include reconstruction-based methods~\cite{xie2022simmim,zhou2019models,he2022masked,chen2023masked,zhou2023self}
and contrastive learning-based methods~\cite{grill2020bootstrap,zbontar2021barlow,chen2021exploring,zhou2023unified,goncharov2023vox2vec}. 
While these methods have shown effectiveness for various downstream vision tasks, they primarily focus on low-level (pixel- or patch-level) image understanding. This focus results in limitations when capturing high-level features, particularly in the semantic interpretability of 3D medical images, restricting the effectiveness of these models in fully utilizing the complex information contained in 3D medical scans. 

Integrating natural language processing and computer vision through language descriptions is an effective way to enhance semantic understanding of images. Recently, multimodal large language models (MLLMs)~\cite{li2023blip,alayrac2022flamingo,2023GPT4VisionSC} have shown remarkable capabilities in enabling natural language to describe and comprehend various visual scenes. Notable examples include BLIP\cite{li2023blip}, Flamingo~\cite{alayrac2022flamingo}, and GPT-4 Vision~\cite{2023GPT4VisionSC}. While MLLMs excel in processing 2D image content, their ability to understand the more complex 3D medical volumes remains an open question. 
Some studies~\cite{chen2023medblip,chen2023generative} explore leveraging existing MLLMs for 3D medical image understanding. Although these methods align image and text features, they fall short in providing the semantic comprehension of 3D images that a dedicated image encoder in an MLLM could achieve. Additionally, these models emphasize a high-level semantic understanding of images while neglecting the low-level details, which are crucial for tasks such as segmentation. 

In summary, a pertaining 3D model that integrates both high-level and low-level semantic understanding is essential for developing comprehensive representations of medical image volumes.
To achieve this, we propose enhancing a 3D image encoder using reconstruction-based self-learning and image-text alignment techniques similar to those employed in MLLMs. 
However, unlike the abundant image-text datasets in general computer vision, datasets containing image-text pairs for 3D medical scans are exceedingly rare, and even collecting a substantial set of 2D medical image-text pairs is a non-trivial task. Furthermore, there is a considerable gap between the level of understanding needed for 3D medical images and the capabilities of current MLLMs designed for 2D natural images. Despite these challenges, GPT-4V(ision)~\cite{2023GPT4VisionSC} has demonstrated some proficiency in interpreting 2D medical images~\cite{yang2023performance,wu2023can}, indicating the potential of leveraging 2D MLLMs for 3D medical image understanding. Therefore, we propose a novel framework, Med3DInsight, which aims to enhance the high-level and low-level image understanding capabilities of an existing 3D image encoder by leveraging 2D MLLMs and reconstruction-based techniques, respectively.

\begin{figure*}[t]
\includegraphics[width=\textwidth]{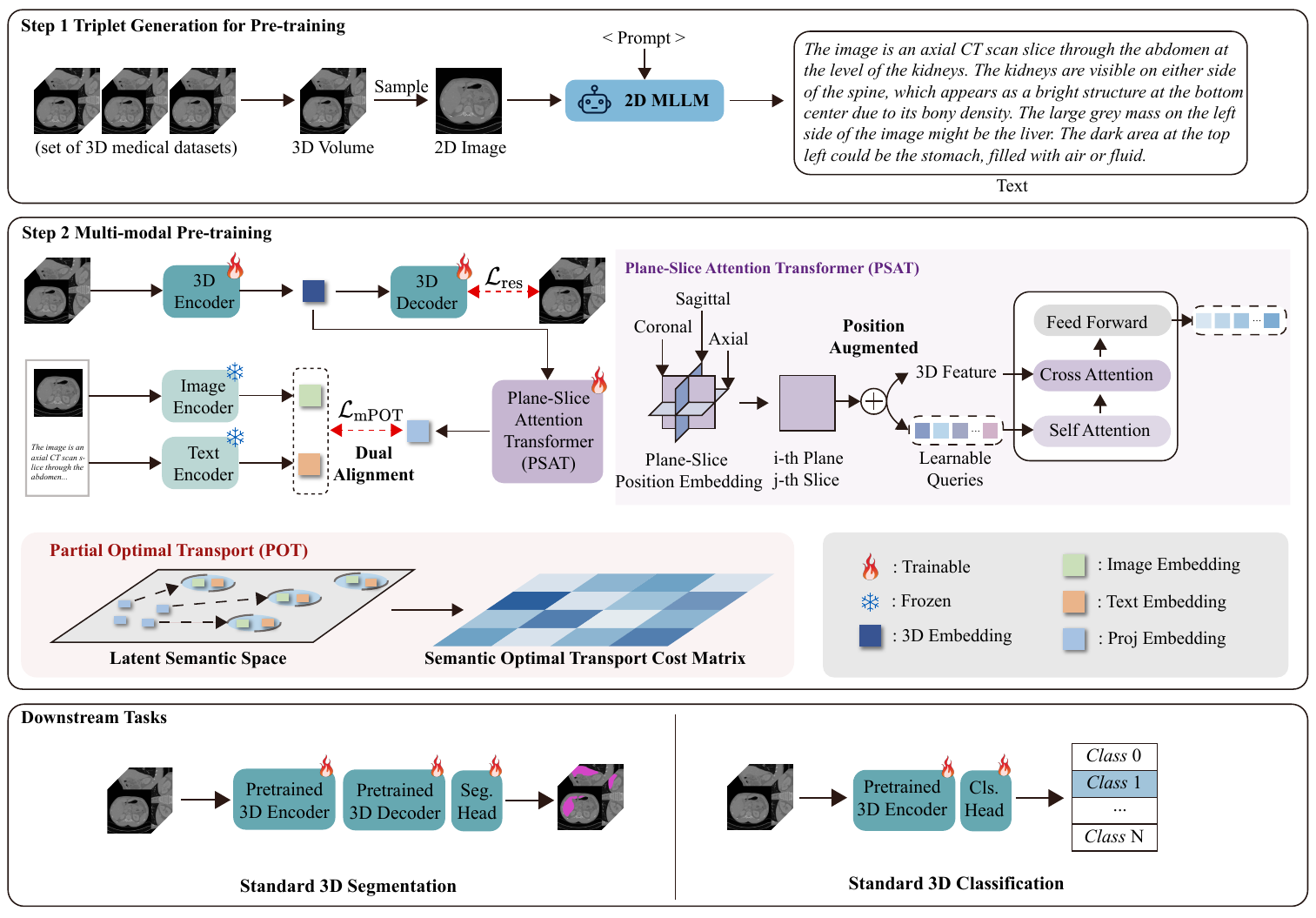}
\caption{Overview of our Med3DInsight framework. During pre-training (\textbf{Top}), we train a 3D image encoder-decoder and a plan-slice-aware transformer (PSAT) that aligns 3D image features with 2D slices and text features extracted from 2D MLLMs.
The pre-trained 3D encoder is further fine-tuned in the downstream tasks, including 3D segmentation (\textbf{Bottom Left}) and 3D classification (\textbf{Bottom Right}).
} 
\label{fig:overview}
\end{figure*}

As illustrated in Fig.~\ref{fig:overview}, Med3DInsight learns 3D medical image representation by leveraging reconstruction and aligning 3D volume features with both 2D image and text features. To bridge the gap between 3D and 2D feature spaces, we introduce a Plane-Slice-Aware Transformer (PSAT) module. This module embeds the plane and slice position within a 3D volume and employs a learnable query technique~\cite{carion2020end} to project 3D features into a 2D feature space for effective alignment and mapping. To generate detailed text descriptions for 2D medical slices extracted from 3D image scans, we utilize GPT-4V(ision)~\cite{2023GPT4VisionSC}, which enhances the model's semantic understanding capability. We then fine-tune image and text encoders of CLIP~\cite{radford2021learning} using the slice-text pairs generated by GPT-4V to align image and text modalities within the same feature space~\cite{xue2023ulip}. 
To address potential misalignment issues between images and text descriptions caused by the noise effects of MLLMs, we employ Partial Optimal Transport (POT) techniques in place of contrastive learning for improved matching accuracy. During pretraining, we keep the GPT-4V and CLIP models frozen while training the 3D image encoder and PSAT module. This approach results in a pre-trained 3D image encoder that is well-prepared for downstream tasks such as segmentation and classification. 

Overall, our contributions are summarized as follows:
\begin{itemize}
    \item We propose a new framework, Med3DInsight, which utilizes 2D MLLMs and reconstruction techniques to enhance both high-level and low-level understanding of an existing 3D image encoder. Our framework is general and achieves state-of-the-art (SOTA) performance in 3D segmentation and classification across multiple datasets.
    \item We design a Plane-Slice-Aware Transformer (PSAT) module to connect the 3D medical image encoder with 2D vision-language models. This module facilitates learning a mapping that considers the spatial orientation of visual features, which is particularly beneficial for applications that require alignment between 2D and 3D feature spaces. 
    \item We propose the Partial Optimal Transport (POT) alignment to further reduce discrepancies across modalities and dimensions, minimizing the noisy effect produced by the potential noise issues in MLLMs. This technique serves as a superior alternative to contrastive learning for self-learning with imperfect paired data. 
\end{itemize}

\section{Related work}
\subsection{Vision Self-supervised Learning}
\subsubsection{Reconstruction-based Methods}
Self-supervised learning (SSL) has significantly advanced computer vision by utilizing pretraining tasks that do not require annotations. A classical technique in SSL involves image restoration to learn visual features by reconstructing corrupted images back to their original state~\cite{wei2022masked,xie2022simmim,he2022masked,gidaris2018unsupervised}.
Some of these approaches have been adapted for medical image analysis to pretrain medical image understanding models. For instance, a context restoration strategy is proposed in~\cite{chen2019self}, which restores randomly swapped image patches to its original version, modified from Masked AutoEncoders (MAE~\cite{he2022masked}). Model Genesis~\cite{zhou2019models} leverages recurrent anatomical patterns in 3D medical images for SSL, thus eliminating the need for manual annotations. Other methods like MIM~\cite{chen2023masked} and medical MAE~\cite{zhou2023self} have also explored masked image modeling for 3D medical image analysis. However, these methods are limited to pixel-level image understanding with local visual semantics. We explore both global and local semantic learning by aligning with 2D MLLMs and reconstructing 3D medical image inputs.

\subsubsection{Alignment-based Methods}
Recently, contrastive learning has emerged as the standard method in SSL~\cite{grill2020bootstrap,zbontar2021barlow}. To address the limitations of traditional contrastive learning, such as the requirement for large batch sizes and strong augmentations~\cite{he2020momentum}, BYOL~\cite{grill2020bootstrap} and BarlowTwins~\cite{zbontar2021barlow} employ a dual-branch structure to align the embeddings of two augmented images, eliminating the need for negative samples in contrastive learning. 
SimSiam~\cite{chen2021exploring} demonstrates the importance of the stop-gradient mechanism on the dual-encoder, introducing a model without negative samples. 
In the context of SSL tailored for medical images, 
PCRLv2~\cite{zhou2023unified} combines contrastive learning with reconstruction tasks, while 
vox2vec~\cite{goncharov2023vox2vec} introduces a fine-grained contrastive learning approach using a Feature Pyramid Network (FPN) to create multi-scale representations.

\subsection{Medical Vision-Language Pretraining}
Medical Vision-Language Pretraining (VLP)~\cite{zhang2022contrastive} has been introduced to integrate textual information into medical image SSL. However, the exploration of medical SSL VLP has been primarily limited to 2D images~\cite{wang2022medclip,huang2021gloria}, 
due to the complexity of medical textural records and the scarcity of large-scale medical image-text datasets.
Nonetheless, 3D medical images, such as MRI, CT, and PET scans, provide more comprehensive and valuable information than their 2D counterparts. Recent research~\cite{butoi2023universeg,ye2023uniseg,wu2023towards,bai2024m3d} 
have extended Med-VLP to include 3D images by using datasets consisting of 3D medical image-text pairs. The lack of publicly available datasets containing large-scale 3D medical image-text pairs restricts the generalizability of VLP to a broader range of medical applications. 

To overcome this limitation, some recent approaches have utilized unpaired data without labels by employing generative methods. For instance, GTGM~\cite{chen2023generative} proposes generating reports from images using a report generator module fine-tuned on MediCaT. Another approach~\cite{liu2023utilizing} suggests using clinical reports exclusively to generate synthetic images. This method employs RoentGen~\cite{chambon2022roentgen}, a pre-trained diffusion model adapted to create synthetic Chest X-rays based on MIMIC-CXR radiology reports. 
The synthetic dataset, including the radiology reports, was used for pretraining a model that achieved comparable results to those trained on the real dataset. 
Additionally, MEDIMP~\cite{milecki2023medimp} utilizes clinical tabular data to generate vocabulary-rich text reports from template sentences and large language models (LLM).

These efforts represent significant advancements in improving SSL methods for 3D medical image analysis. 
However, our work stands out due to two key innovations: (1) incorporating generative description as a self-supervised signal, and leveraging clinical domain knowledge from advanced 2D MLLMs to alleviate the major challenge of lacking large-scale image-text pairs in medical VLP, and (2) developing a self-supervised learning framework that seamlessly integrates high-level semantic learning with low-level restorative learning and effectively bridges the gap between 3D image features with 2D image and text features. 

\subsection{Optimal Transport for Alignment}

Optimal transport (OT) is a powerful mathematical tool to measure the discrepancy between two probability distributions. 
Therefore, it has a wide range of applications, such as domain adaption~\cite{courty2017joint,chen2024disentangle,wang2024spatial} and representation learning~\cite{chen2020uniter,luo2022weakly,wang2023self,dong2023partial,zhang2024prototype}.
Our work is close to the Inverse Optimal Transport (IOT) problem, which has been well-studied~\cite{li2019learning,stuart2020inverse,luo2022weakly,wang2023self} 
by the research community aiming to learn matching between two sources of interest. 
For instance, the Vision-Language Pretrained (VLP) model, Uniter~\cite{chen2020uniter}, applies optimal transport to promote fine-grained alignment between words and image regions, resulting in better joint embeddings for downstream tasks. 
Researchers~\cite{wu2021data} proposed the OTTER framework which leverages entropic optimal plan as soft matching labels to learn many-to-many relationships in the image-text matching task based on contrastive learning procedure.
These applications demonstrate the remarkable versatility of inverse optimal transport in cross-modal alignment and inspire us to develop an IOT-based framework for medical understanding.
Most existing methods solve optimal transport problems in the Kantorovich form (i.e., Wasserstein distance)~\cite{benamou2000computational}. Some efficient approximate algorithms have been proposed, such as Sinkhorn scaling algorithm~\cite{cuturi2013sinkhorn}. 
The work in~\cite{luong2024revisiting} adapt the alignment framework with a more flexible ground cost metric, Mahalanobis distance~\cite{de2000mahalanobis}, to model the expressive shared embedding space between audio and text modalities, which obtains encouraging performance in audio-text retrival. 
In this study, we propose a new IOT-based alignment method for self-supervised 3D medical understanding. Moreover, to cope with misaligned training data in practice, we propose a variant using partial optimal transport to mitigate the harm of misaligned data pairs in training data.

\section{Methodology}
Figure~\ref{fig:overview} presents our Med3DInsight framework, which comprises two main phases: generating a triplet dataset using 2D MLLMs and pretraining our model through 2D/3D alignment and 3D reconstruction.

\subsection{Triplet Generation: 3D Image Volume, 2D Image Slice and Corresponding Textual Descriptions}
\label{sec:triplets_gen}

\begin{table}[t]
\begin{center}
\begin{threeparttable} 
\caption{The information for pretraining datasets.}
\label{tab:pretrain_datasets}
\begin{tabular}{p{50pt}p{160pt}} 
\toprule
Modalities & MRI (4,587), CT (18,330) \\
\cmidrule(r){1-2}
Planes & Axial (13,034), Coronal (3,860), Sagittal (7,242) \\
\cmidrule(r){1-2}
Body Parts & Head (8,654), Chest (3,846), Abdomen (6,093) \\
\cmidrule(r){1-2}
Internal Organs & Brain (9,446), Heart (4,604), Liver (6,684), Stomach (1,333), Vessel (2,415), Spleen (2,732), Vertebral (4,678), Kidney (4,862) \\
\cmidrule(r){1-2}
Target Types & Tissue (8,291), Organ (10,271), Tumor (837), Lesion (469) \\ 
\bottomrule
\end{tabular}
\begin{tablenotes} 
\small
\item[*] Values in parentheses indicate the number of associated data triplets.
\end{tablenotes}
\end{threeparttable}
\end{center}
\end{table}

During the generation process, we work with a set of $N$ 3D medical image volumes denoted as $X = \{x_1, x_2, ..., x_N \}$. For each 3D volume $x_k$, we randomly select one 2D image slice $s_k^{(i,j)}$, which is extracted from the $i$-th plane (i.e., coronal, sagittal, or axial) and the $j$-th slice. To generate textual descriptions for such 2D image slices, we employ an MLLM generator $G(\theta)$, specifically GPT-4V~\cite{2023GPT4VisionSC}, which has shown effectiveness in describing 2D medical scans~\cite{wu2023can,yang2023performance}. The GPT-4V prompting protocol we used is simply: ``Describe the image in fewer than 100 words".
The selected 2D slice $s_k^{(i,j)}$ serves as input to the text generator $G(\theta)$, which then generates its textual description $t_k^{(i,j)}$ to form a triplet $D_k:(x_k, s_k^{(i,j)}, t_k^{(i,j)})$. In this way, we collect our triplet dataset $D = \{(x_1, s_1^{(i_i,j_1)}, t_1^{(i_1,j_1)}), (x_2, s_2^{(i_2,j_2)}, t_2^{(i_2,j_2)}), \cdots, (x_N, s_N^{(i_N,j_N)}, $ $t_N^{(i_N,j_N)})\}$, which consists of 24,140 triplets and is utilized in the subsequent pretraining step~\footnote{Both the triplet dataset and our pre-trained models are publicly released at \href{https://github.com/Qybc/Med3DInsight}{https://github.com/Qybc/Med3DInsight}.}.

For this study, we construct the slice-text triplet dataset using two publicly available 3D medical image datasets: 3DSeg-8~\cite{chen2019med3d} and M3D~\cite{bai2024m3d}. These datasets provide a strong foundation for ensuring diversity in imaging modality, anatomical coverage, and pathology types. As summarized in Table~\ref{tab:pretrain_datasets}, we perform a distributional analysis to assess the balance across imaging modalities, body regions, and clinical conditions. These pretraining datasets offer three key advantages: (1) Multi-modality representation: The inclusion of both CT and MRI data enables the model to learn from heterogeneous imaging characteristics. For instance, CT captures tissue density and MRI emphasizes contrast variations.
(2) Comprehensive anatomical coverage: The datasets span a wide range of anatomical regions from head to abdomen, including 8,654 head scans, 3,846 chest scans, and 6,093 abdominal scans.
(3) Clinically diverse pathology types: A broad spectrum of diagnostic targets is included, such as tissue (8,291 samples), organs (10,271), tumors (837), and lesions (469), supporting robust semantic learning.

To construct slice-text triplets for pretraining, we randomly sample a single 2D slice from each 3D volume. Although each volume contains numerous slices, this design choice is made for two primary reasons:
(1) Mitigating redundancy: Adjacent slices within a volume often exhibit minimal anatomical variation (e.g., sequential axial liver scans), leading to highly similar visual content and repetitive textual descriptions. Moreover, even across different subjects, slices from the same anatomical region tend to share structural similarity, further reducing the marginal utility of additional samples. Including such redundant slices would bias the model toward overrepresented anatomies while inflating computational cost.
(2) Promoting diversity: We select one slice per volume across a wide range of modalities, anatomical regions, and pathological conditions, which maximizes heterogeneity in the pretraining data. This diversity is crucial for improving the model’s generalization across unseen clinical scenarios. Representative examples of the generated slice-text pairs are shown in Table~\ref{tab:slice_and_description}. Notably, all triplets are constructed without using any manual annotations or diagnostic labels.

\begin{table*}[htbp]
\scriptsize
\centering
\caption{Samples of the textual descriptions generated for 2D slices. (\textcolor{blue}{Blue}: accurate description; \textcolor{red}{red}: inaccurate).}
\label{tab:slice_and_description}
\begin{tabular}{p{10pt}p{45pt}p{380pt}}
\cmidrule{1-3}
& Slices & Descriptions \\
\cmidrule{1-3}
(a) & \begin{minipage}[b]{0.18\columnwidth} \raisebox{-.9\height}{\includegraphics[width=\linewidth]{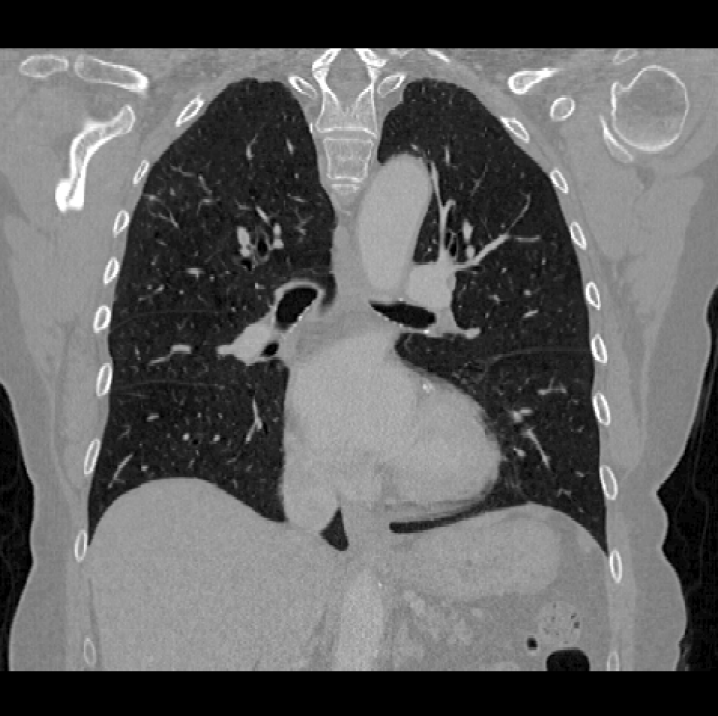}} \end{minipage} & This image is \cBlue{\textbf{a coronal CT scan of the chest, displaying the lungs, heart, and surrounding anatomical structures. The lungs appear as large, dark areas on either side of the image, with visible branching airways. The heart is centrally located, appearing as a lighter, more solid structure. The spine is visible in the background, running vertically along the center. The ribs are seen as curved, lighter lines on either side of the chest. There are also various soft tissues and blood vessels visible.}} The scan provides detailed cross-sectional imaging, useful for diagnosing and evaluating thoracic conditions. \\
\cmidrule{1-3}
(b) &  \begin{minipage}[b]{0.18\columnwidth} \raisebox{-.9\height}{\includegraphics[width=\linewidth]{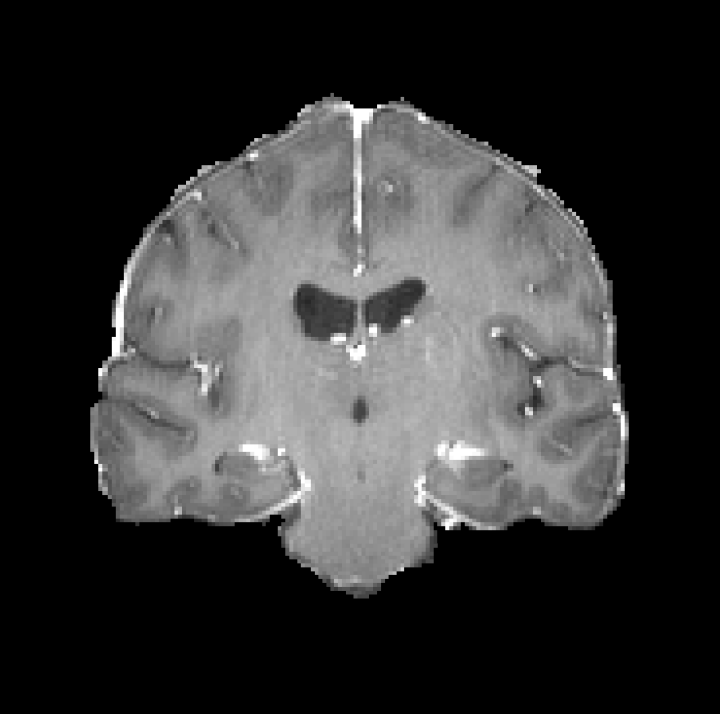}} \end{minipage}  & This image is \cBlue{\textbf{a coronal MRI scan of the brain, providing a crosssectional view of the cerebral structures. The scan clearly shows the symmetrical arrangement of the brain's hemispheres, with the central sulcus and other gyri and sulci visible on the surface. The ventricular system is also apparent, with the lateral ventricles appearing as dark, butterfly-shaped structures in the middle. The surrounding gray matter and white matter can be distinguished by their varying shades of gray.}} This type of imaging is crucial for diagnosing and evaluating neurological conditions, as it highlights differences in tissue density and structure within the brain. \\
\cmidrule{1-3}
(c) &  \begin{minipage}[b]{0.18\columnwidth} \raisebox{-.9\height}{\includegraphics[width=\linewidth]{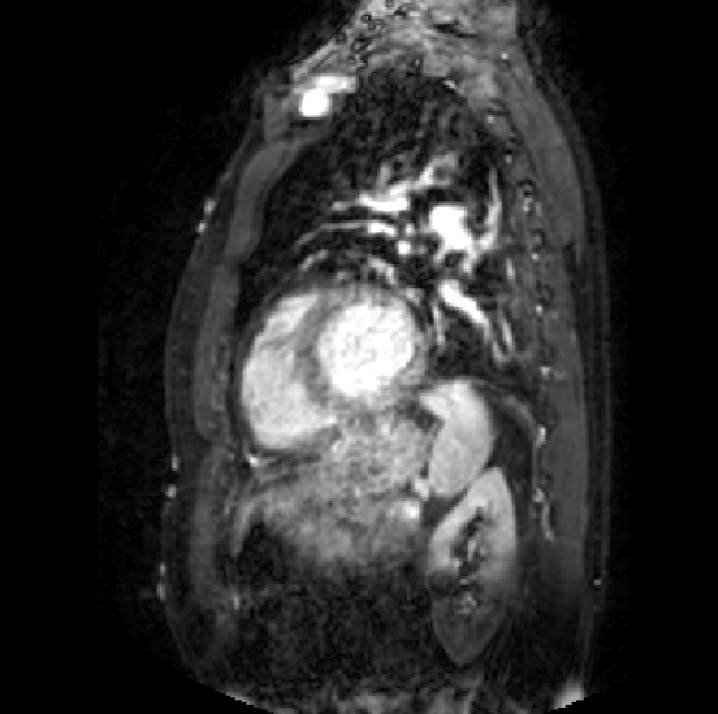}} \end{minipage}  & This image is \cBlue{\textbf{a sagittal MRI scan of the thoracic region, prominently displaying the heart and surrounding structures. The heart is centrally located, with its chambers and major blood vessels visible, including the aorta. The lungs, with their branching bronchial structures, can be seen around the heart. The vertebral column runs vertically along the left side of the image, providing structural support and orientation.}} The grayscale contrast highlights different tissue densities, aiding in distinguishing between various anatomical features. This detailed imaging is crucial for diagnosing cardiovascular and thoracic conditions.\\
\cmidrule{1-3}
(d) &  \begin{minipage}[b]{0.18\columnwidth} \raisebox{-.9\height}{\includegraphics[width=\linewidth]{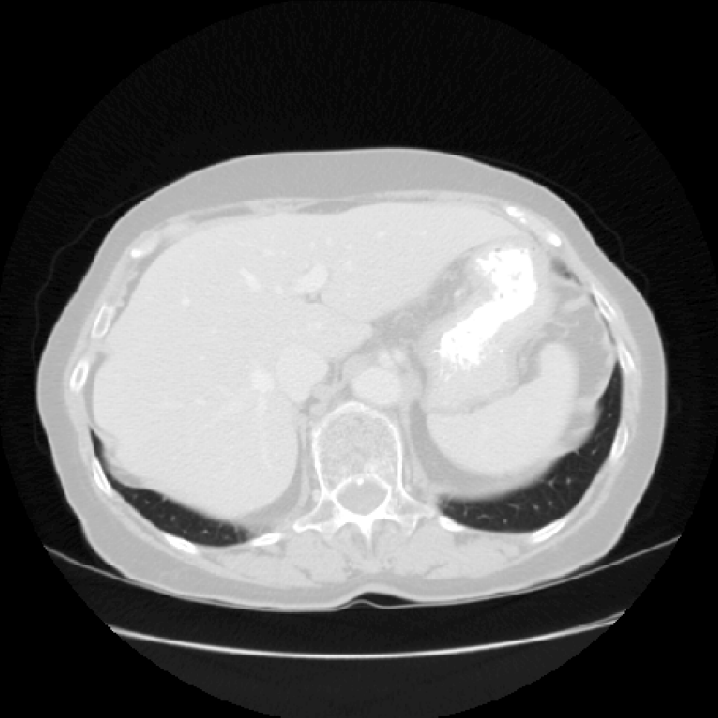}} \end{minipage}  & This image is \cBlue{\textbf{a transverse CT scan of the abdomen, offering a detailed cross-sectional view of the internal organs. The large, rounded structure in the center is likely the liver, with its distinct texture and density. Adjacent to it, the stomach and spleen are visible, with the spleen showing a brighter area indicative of different tissue or a lesion. The vertebral column is centrally located at the bottom, providing a structural reference.}} The grayscale contrast differentiates between various tissue densities, essential for identifying abnormalities. This detailed imaging is crucial for diagnosing abdominal conditions and guiding further medical evaluation and treatment.\\
\cmidrule{1-3}
(e) &  \begin{minipage}[b]{0.18\columnwidth} \raisebox{-.9\height}{\includegraphics[width=\linewidth]{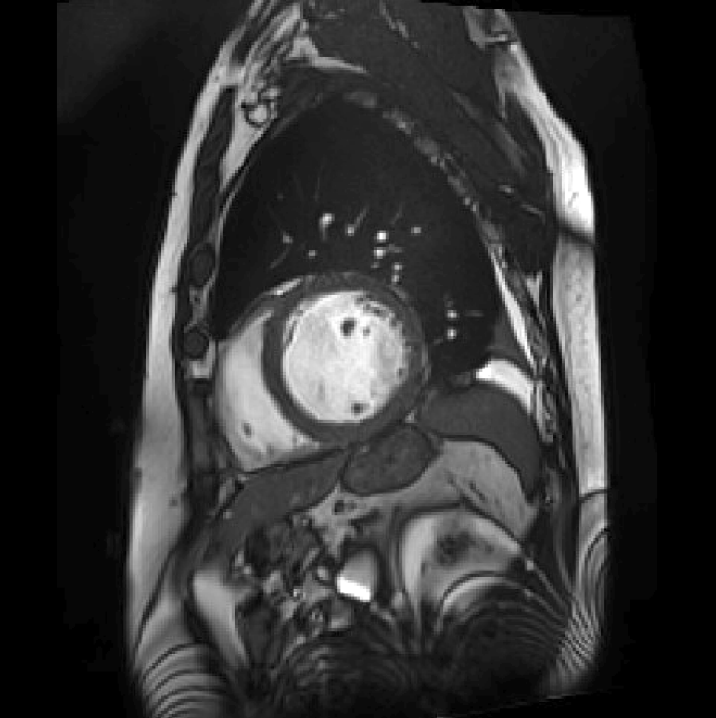}} \end{minipage}  & This image is \cBlue{\textbf{a sagittal MRI scan}} of the \cRed{\textbf{abdomen}}, showcasing various internal structures in detail. \cRed{\textbf{Central to the image is a large, circular organ, likely the liver, with its characteristic texture and surrounding tissues.}} \cBlue{\textbf{The diaphragm is visible, delineating the boundary between the thoracic and abdominal cavities.}} \cBlue{\textbf{The vertebral column is seen running vertically along the left side, providing a reference point for orientation.}} The grayscale contrast highlights different tissue densities, facilitating the diagnosis and evaluation of abdominal conditions. \\
\cmidrule{1-3}

\end{tabular}
\end{table*}

\subsection{Triplet Representation Learning}
Our framework employs an encoder-decoder structure as its 3D visual backbone and ensures alignment with 2D image and text features via pre-trained and pre-aligned image and text encoders. 
For 2D image and text encoders, we prefer those that are pre-aligned in existing vision-language models, e.g., CLIP~\cite{radford2021learning}. Detailed settings are provided in Section~\ref{sec:settings}.

Specifically, given the dataset $D$ consisting of 3D medical volumes paired with 2D image-text pairs, our goal is to learn a 3D visual representation using a 3D volume encoder $f_V (\cdot)$ with the help of a pair of pre-aligned 2D image encoder $f_I (\cdot)$ and text encoder $f_T (\cdot)$. 
For instance, given a sample triplet $D_k:(x_k, s_k^{(i,j)}, t_k^{(i,j)})$, we compute their respective feature representations: 
$h^x_{k} =  f_V (x_k)$ for the 3D volume $x_k$,
$h^s_{(i,j,k)} =  f_I (s_k^{(i,j)})$ for the 2D image slice $s_k^{(i,j)}$, and
$h^t_{(i,j,k)} =  f_T (t_k^{(i,j)})$ for the textual description $t_{(i,j,k)}$. Thus, for each triplet $D_k$, we obtain its triplet feature set $H_k:(h^x_k, h^s_{(i,j,k)}, h^t_{(i,j,k)})$. This feature set $H_k$ encapsulates the encoded representations of the 3D medical volume, a randomly selected 2D image slice, and its corresponding textual description, respectively.

\subsection{Triplet Alignment}
In the triplets $\{H_k\}$, the text features are pre-aligned with the image features of their corresponding 2D slices. However, a disparity exists between the 3D image features $\{h^v_k\}$ and the pairs of 2D image and text features $\{(h^s_{(i,j,k)}, h^t_{(i,j,k)})\}$. Bridging this gap between 2D and 3D feature representations is crucial for enabling effective self-learning during the pretraining stage. One solution is to align 3D image features with the features of image slices and text individually. 

To achieve alignment, we employ a query techinque, which is further enhanced by incorporating knowledge of the plane-slice location of the randomly selected image slice. As depicted in Fig.~\ref{fig:overview}, we propose a Plane-Slice Attention Transformer (PSAT) module, which adopts a collection of $Q$ learnable query embeddings to facilitate a projection of 3D volume features into the 2D image and text embedding space of CLIP.   
These queries serve as adaptive parameters that learn to decode 3D volume features into slice-specific 2D representations. Each query functions as a spatial probe, selectively attending to regions within the 3D volume that are anatomically relevant to the target 2D slice.
Specifically, these learnable queries interact with each other through a self-attention layer, then engage with the volume feature $h^v_k$ via a cross-attention layer, producing an output of $Q$ encoded visual vectors, one for each query embedding.
These vectors are subsequently processed by a multi-layer perceptron (MLP), resulting in the projection of the volume embedding $\{\hat{h}^v_k\}$. This projected embedding is then aligned with the image slice and text embeddings using an OT-based alignment method, as described in Section~\ref{sec:iot}.

To inform the location of an image slice within its associated image volume, we propose incorporating the actual slice position into the query learning process. Specifically, the image slice $s_k^{(i,j)}$ is extracted from the image volume $x_k$ and is located at the $j$-th slice of the $i$-th plane, the 3D-2D projection should be aware of this orientation relationship between image volume and slice. Therefore, we introduce a plane-slice position embedding for the volume feature to promote the model’s ability to learn orientation and geometric relationships. Specifically, we first construct the plane-slice position embedding $V_p\in \mathbb{R}^{C \times I \times J}$ with zero-initialized parameters, where $C$ is the embedding dimension of the model, $I$ is the number of planes, and $J$ is the number of slices.
During pre-training, when dealing with an image slice $s_k^{(i,j)}$, for instance, when a sample corresponds to the eighth slice (i.e., $j=8$) of the coronal plane (i.e., $i=0$), we inject the plane-slice position embedding $V_p\in \mathbb{R}^{c \times 1 \times 1}$ for the eighth slice at the coronal plane into the volume feature and queries, following general positional encoding in~\cite{vaswani2017attention}.

By incorporating actual slice position into the query learning process, the model gains precise spatial awareness of each slice’s anatomical context within the 3D volume. This spatial grounding allows queries to act as informed selectors rather than generic decoders, significantly improving the fidelity of 2D representations derived from volumetric data. Such alignment is critical for leveraging powerful pre-trained 2D MLLMs in 3D medical image understanding. The improved 3D-to-2D projection serves as a robust bridge between volumetric data and text supervision, enabling effective pretraining without the need for manual annotations.

\subsection{Learning Objectives}

\subsubsection{Alignment Loss}
\label{sec:iot}

Contrastive learning, as used in CLIP~\cite{radford2021learning}, offers an efficient self-learning technique by bringing the representations of paired data closer together in the embedding space while pushing apart the representations of non-paired data. However, the use of large language models inevitably leads to some noise issues, as shown in Table~\ref{tab:slice_and_description}(e), resulting in text descriptions that are imperfectly paired with image slices. Also, 3D image features are partially paired with those of 2D slices and text descriptions. These are challenges faced by standard contrastive learning techniques.

Optimal Transport (OT)~\cite{peyre2019computational} studies how to move and distribute resources or mass and find the optimal way to transport mass from one distribution to another while minimizing a certain cost. When the source and target distributions are unequal, partial optimal transport (POT)~\cite{chizat2018scaling} is a feasible solution. To address misaligned or partially aligned triplet features, we propose a variant of POT to mitigate their impact during pretraining. This approach allows for partial alignment, where only a subset of the 3D features are aligned with the 2D image and text features, demonstrating improved noise tolerance.
Specifically, due to the presence of noisy correspondences, the empirical matching $\hat{\pi}$ of POT becomes corrupted compared to standard OT, leading to certain rows without a match.

We denote the incomplete optimal transport plan  $\pi_{s, \epsilon, c}^{X, Y}$ as:
\begin{equation}
    \pi_{s, \epsilon, c}^{X, Y}=\underset{\pi \in \Pi_{s}\left(P_{X}, P_{Y}\right)}{\arg \min } \sum_{i=1}^{n} \sum_{j=1}^{n} \pi_{i j} c\left(x_{i}, y_{j}\right)-\epsilon \sum_{i=1}^{n} \sum_{j=1}^{n} \pi_{i j} \log \pi_{i j} ,
    \label{eq:ot}
\end{equation}
where $\Pi_{s}\left(P_{X}, P_{Y}\right)$ denotes the set of incomplete transportation
plans or couplings between two empirical probability measures $P_{X}$ and $P_{Y}$, $s$ is the amount of mass to be transported, $x_i$ and $y_j$ are the pair of two samples, $\pi_{ij}$ is the coupling value of $x_{i}$ and $y_{j}$, and $c$ is the ground metric, e.g., Euclidean distance or Mahalanobis distance. When $\epsilon > 0$, the optimization problem can be
solved by using a generalization of the Bregman projections algorithm~\cite{benamou2015iterative}.
Therefore, given a mini-batch $(X^b, S^b, T^b) \sim D$, the mini-batch POT (mPOT) solves the following optimization problem:
\begin{equation}
    \mathcal{L}_{\text{mPOT}} = \mathbb{E}_{(X_b, S_b, T_b) \sim D}[KL(\hat{\pi}^{b}||\pi_{s,\epsilon,c}^{X^{b},S^{b}}) + KL(\hat{\pi}^{b}||\pi_{s,\epsilon,c}^{X^{b},T^{b}})]. \label{eq:mPOT_loss}
\end{equation}
Here, $b\geq 2$ denotes the mini-batch size, $\pi_{s,\epsilon,c}^{X^{b},S^{b}}$ and $\pi_{s,\epsilon,c}^{X^{b},T^{b}}$ represent the transportation plan. These plans aim to align volume features with slice features and text features, respectively. Additionally, $\hat{\pi}^{b}$ is a $b \times b$ matrix such that $\hat{\pi}_{ii} = 1/b$ and $\hat{\pi}_{ij} = 0 (i \neq j)$, ensuring uniformity within the mini-batch.
Given that embeddings from encoders may lack proper alignment in terms of scaling across dimensions~\cite{cuturi2014ground}, we tackle this issue by substituting the commonly used Euclidean distance~\cite{danielsson1980euclidean} with the Mahalanobis distance~\cite{dhouib2020swiss} as the ground metric. 
For more information, readers can refer to the training algorithm presented in Algorithm~\ref{alg:algorithm1}.

\begin{algorithm}[t]
	\caption{Learning mini-batch partial optimal transport (mPOT) cost metric using Mahanalobis distance.}
	\label{alg:algorithm1}
	\KwIn{Initialized 3D volume encoder $f_{V(\theta)}$ and 3D volume projection network $f_{Proj(\psi)}$, frozen 2D image encoder $f_I$ and text encoder $f_T$, initialized interaction matrix $M$ in the Mahalanobis distance, training data $D$, batch size $b$, amount of mass to be transported $s$, entropy regularization coefficient $\epsilon$ and learning rate $\alpha$}
	\KwOut{Learned 3D volume encoder $f_{V(\theta)}$, projection network $f_{Proj(\psi)}$, and interaction matrix $M$}  
	\BlankLine
	\While{not converged}{
        Draw a mini-batch from the training set: $(X^b, S^b, T^b) = (x_{i}, s_{i}, t_{i})_{i=1}^{b} \sim D;$ \\ 
        Extract projected volume embedding, image slice embedding, and text embedding, respectively: \\
        \quad $\hat{H}^{X^b} = \{\hat{h}_i^x\}_{i=1}^b=\{f_{Proj(\psi)}(f_{V(\theta)}(x_{i}))\}_{i=1}^{b},$ \\
        \quad $H^{S^b} = h_i^s=\{f_{I}(s_{i})\}_{i=1}^{b},$ \\
        \quad $H^{T^b} = h_i^t=\{f_{T}(t_{i})\}_{i=1}^{b};$ \\
        Create a $b\times b$ matrix with diagonal values set to $1/b$:
        $\hat{\pi}=\operatorname{eye}(b, \frac{1}{b}) ;$ \\
        Calculate the Mahalanobis distance matrix within the mini-batch: \\
        \quad (a) between volume and slice embeddings: \\
        $C_{M}^{\hat{H}^{X^b}, H^{S^b}}=\sqrt{(\hat{H}^{X^b}-H^{S^b})^{\top} M(\hat{H}^{X^b}-H^{S^b})} ,$ \\
        \quad (b) between volume and text embeddings: \\
        $C_{M}^{\hat{H}^{X^b}, H^{T^b}}=\sqrt{(\hat{H}^{X^b}-H^{T^b})^{\top} M(\hat{H}^{X^b}-H^{T^b})};$\\
        Compute the transportation plans using Bregman~\cite{benamou2015iterative} to match volume with slice and text, respectively:\\
        \quad $\pi_{s, \epsilon, C(\theta, \psi, M)}^{X^{b}, S^{b}}=\operatorname{Bregman}(s, C_{M}^{\hat{H}^{X^b}, H^{S^b}}, \epsilon) ,$ \\
        \quad $\pi_{s, \epsilon, C(\theta, \psi, M)}^{X^{b}, T^{b}}=\operatorname{Bregman}(s, C_{M}^{\hat{H}^{X^b}, H^{T^b}}, \epsilon) ;$ \\ 
        Estimate the mPOT loss function in Eq.~\ref{eq:mPOT_loss}: \\
        \quad $\mathcal{L}(\theta, \psi, M)=KL(\hat{\pi} \| \pi_{s, \epsilon, C(\theta, \psi, M)}^{X^{b}, S^{b}}) + KL(\hat{\pi} \| \pi_{s, \epsilon, C(\theta, \psi, M)}^{X^{b}, T^{b}}) ;$ \\
        Update the learnable parameters: \\
        \quad $\theta=\theta-\alpha \nabla_{\theta} \mathcal{L} ,$ \\
        \quad $\psi=\psi-\alpha \nabla_{\psi} \mathcal{L} ,$ \\
        \quad $M=\operatorname{Proj}(M-\alpha \nabla_{M} \mathcal{L}) .$ \\
        }
	\Return{$f_{V(\theta)}$, $f_{Proj(\psi)}$, and $M$}
\end{algorithm}

\subsubsection{Reconstruction Loss}
To enhance visual feature learning with low-level visual semantics, we introduce an auxiliary pretext task, image restoration, which aims to reconstruct the 3D medical volume using latent visual encoders. The training objective for this task is formulated using the L2-norm below:
\begin{equation}
    \mathcal{L}_{\text{res}} = \mathbb{E}_{(X^{b})} \sum_{i=1}^{b} ||x_i - \hat{x}_i||_{2} . 
\end{equation}

\subsubsection{Total Loss}
With a synergy of triplet alignment learning and image reconstruction, the overall pretraining objective of our Med3DInsight framework is represented as follows:
\begin{equation}
    \mathcal{L}_{\text{total}} = \mathcal{L}_{\text{mPOT}} + \mathcal{L}_{\text{res}}
\end{equation}
After pretraining, the resulting 3D image encoder serves as a visual backbone for downstream task-specific fine-tuning.

\section{Experiments}

\subsection{Pretraining Setup and Settings}
\label{sec:settings}
In Med3DInsight, the nnFormer~\cite{zhou2023nnformer} serves as the 3D visual encoder, and the CLIP~\cite{radford2021learning} serves as the 2D image and text encoder.
We freeze the image and text encoder during pretraining to optimize computational efficiency while extracting pre-aligned embeddings from image slices and texts with CLIP, compelling the 3D visual backbone to learn more representative features. 
However, since the image and generated descriptions are in the medical domain while CLIP has been pre-trained on natural image-text pairs, we fine-tune the CLIP encoders using our generated 2D image-text pairs before our pretraining.
For the PSAT module, we follow a common setting and set the token number of learnable queries $Q$ to 300, and the dimension of the token is 512.
During pretraining, we employ a batch size of 4 on each GPU and utilize the Adam optimizer with an initial learning rate of 2e-5. The training is set for 300 epochs to ensure convergence. All pretraining experiments are implemented on four NVIDIA GeForce RTX 3090 GPUs.

\subsection{Downstream Tasks and Corresponding Datasets}

We evaluate the effectiveness of our method across a diverse set of 3D medical imaging tasks, varying in size, objectives, and modalities.
Specifically, we evaluate the representativity and adaptability of the visual features informed by MLLMs.
These evaluations span four types of distinct 3D medical image segmentation tasks and two 3D medical image classification task, covering over 19 organs or tumors, and spanning both CT and MRI modalities, as listed in Table~\ref{tab:datasets}.

\begin{table}[t]
\scriptsize
\begin{center}
\caption{The information for downstream datasets.
}
\label{tab:datasets}
\begin{tabular}{l@{\hspace{2pt}}c@{\hspace{2pt}}c@{\hspace{2pt}}c@{\hspace{2pt}}l} 
\toprule
Datasets   & Modality & \#Targets & \#Scans & \#Annotated Organs or Tumors\\
\cmidrule(r){1-5}
MM-WHS~\cite{zhuang2016multi} & CT & 7 & 20 & \makecell[l]{
Ventricle Blood Cavity, Right \\ Ventricle Blood 
Cavity, Left Atrium \\ Blood Cavity, Right Atrium 
Blood \\ Cavity, Myocardium of Left Ventricle, \\
Ascending Aorta, Pulmonary Artery}  \\
MSD-Heart~\cite{antonelli2022medical} & T1 MRI & 1 & 20 & Heart   \\
MSD-Liver~\cite{antonelli2022medical} & CT & 3 & 131 & Liver, Kidneys, Spleen \\
MSD-Colon~\cite{antonelli2022medical} & CT & 1 & 126 & Colon cancer \\
CHAOS~\cite{kavur2021chaos} & CT & 1 & 20 & Liver  \\
AbdomenCT-1K~\cite{ma2021abdomenct} & CT & 4 & 361 & Spleen, Kidney, Liver, Pancreas \\
OASIS1~\cite{marcus2007open} & T1 MRI & 4 & 414 & \makecell[l]{Cortex, Subcortical Gray Matter, \\ White Matter, CSF} \\
ADNI~\cite{mueller2005alzheimer} & T1 MRI & 4 & 5069 & \makecell[l]{Cortex, Subcortical Gray Matter, \\ White Matter, CSF} \\
VS~\cite{shapey2021segmentation} & T2 MRI & 1 & 242 & Vestibular Schwannoma Tumor \\
LiTs~\cite{bilic2023liver} & CT & 2 & 131 & Liver, Liver Tumor  \\
\cmidrule(r){1-5}
Datasets   & Modality & \#Classes & \#Scans & \#Annotated Labels\\
\cmidrule(r){1-5}
OASIS2~\cite{marcus2010open} & T1 MRI & 2 & 312 & Demented, Non-demented \\
PPMI~\cite{marek2011parkinson} & T1 MRI & 2 & 582 & Normal Control, PD \\
\bottomrule
\end{tabular}
\end{center}
\end{table}

\begin{table*}[t]
\caption{Comparison with supervised and SSL baselines in Dice(\%) and HD95(mm) on eight 3D segmentation datasets. Our method consistently performs well over the SOTA approaches. The best results are bolded, and the second best are underlined.}
\centering
\label{tab:seg}
\begin{minipage}{0.32\textwidth}
\subcaption{\textbf{Average over 10 datasets}} 
\centering
\scriptsize
\makeatletter\def\@captype{table}\makeatother
\label{seg:avg}
\begin{tabular}{lcc}
\toprule
& \textbf{DSC(\%)$\uparrow$} & \textbf{HD95(mm)$\downarrow$} \\
\cmidrule(r){1-3}
UNet~\cite{ronneberger2015u}  & 80.83 $\pm$ 0.63 & 4.53 $\pm$ 0.86 \\
nnUNet~\cite{isensee2021nnu}  & 83.31 $\pm$ 0.54 & 3.61 $\pm$ 0.71 \\
SwinUNETR~\cite{hatamizadeh2021swin}  & 83.44 $\pm$ 0.55 & 3.46 $\pm$ 0.64 \\
nnFormer~\cite{zhou2023nnformer}  & 84.15 $\pm$ 0.54 & 3.74 $\pm$ 0.58 \\
\cmidrule(r){1-3}
MIM~\cite{chen2023masked} & 84.40 $\pm$ 0.51 & 3.92 $\pm$ 0.57 \\
MedMAE~\cite{zhou2023self} & 85.81 $\pm$ 0.38 & 3.21 $\pm$ 0.46 \\
RPC-Loc~\cite{lei2021contrastive} & 82.44 $\pm$ 0.48 & 4.26 $\pm$ 0.59 \\
VoCo~\cite{wu2024voco} & 86.06 $\pm$ 0.43 & 3.58 $\pm$ 0.49 \\
PCRLv2~\cite{zhou2023unified} & 85.95 $\pm$ 0.39 & 3.75 $\pm$ 0.53 \\
vox2vec~\cite{goncharov2023vox2vec} & \underline{87.52 $\pm$ 0.49} & \underline{2.96 $\pm$ 0.45} \\
\cmidrule(r){1-3}
\rowcolor{myblue}
ours & \textbf{88.59 $\pm$ 0.40} & \textbf{2.49 $\pm$ 0.33} \\
\bottomrule
\end{tabular} 
\end{minipage}\quad
\begin{minipage}{0.32\textwidth}
\subcaption{MM-WHS} 
\centering
\scriptsize
\makeatletter\def\@captype{table}\makeatother
\label{seg:mm-whs}
\begin{tabular}{lcc}
\toprule
& \textbf{DSC(\%)$\uparrow$} & \textbf{HD95(mm)$\downarrow$} \\
\cmidrule(r){1-3}
UNet~\cite{ronneberger2015u}  & 82.16 $\pm$ 0.22 & 2.48 $\pm$ 1.05 \\
nnUNet~\cite{isensee2021nnu}  & 85.36 $\pm$ 0.57 & 3.87 $\pm$ 0.77 \\
SwinUNETR~\cite{hatamizadeh2021swin}  & 85.07 $\pm$ 0.87 & 2.33 $\pm$ 0.42 \\
nnFormer~\cite{zhou2023nnformer}  & 85.86 $\pm$ 0.59 & 2.58 $\pm$ 0.25 \\
\cmidrule(r){1-3}
MIM~\cite{chen2023masked} & 84.12 $\pm$ 0.60 & 3.36 $\pm$ 0.33 \\
MedMAE~\cite{zhou2023self} & 86.02 $\pm$ 0.50 & \underline{2.12 $\pm$ 0.13} \\
RPC-Loc~\cite{lei2021contrastive} & 81.24 $\pm$ 0.33 & 3.22 $\pm$ 0.73\\
VoCo~\cite{wu2024voco} & 86.21 $\pm$ 0.43 & 2.24 $\pm$ 0.33\\
PCRLv2~\cite{zhou2023unified} & 85.58 $\pm$ 0.45 & 2.62 $\pm$ 0.17 \\
vox2vec~\cite{goncharov2023vox2vec} & \underline{87.74 $\pm$ 0.38} & 2.14 $\pm$ 0.28 \\
\cmidrule(r){1-3}
\rowcolor{myblue}
ours & \textbf{88.67 $\pm$ 0.04} & \textbf{1.68 $\pm$ 0.10} \\
\bottomrule
\end{tabular} 
\end{minipage}\quad
\begin{minipage}{0.32\textwidth}
\subcaption{CHAOS}
\centering
\scriptsize
\makeatletter\def\@captype{table}\makeatother
\label{seg:chaos}
\begin{tabular}{lcc}
\toprule
& \textbf{DSC(\%)$\uparrow$} & \textbf{HD95(mm)$\downarrow$} \\
\cmidrule(r){1-3}
UNet~\cite{ronneberger2015u}  & 90.67 $\pm$ 0.83 & 3.01 $\pm$ 0.92 \\
nnUNet~\cite{isensee2021nnu}  & 91.20 $\pm$ 0.18 & 2.17 $\pm$ 0.23 \\
SwinUNETR~\cite{hatamizadeh2021swin}  & 90.34 $\pm$ 0.31 & 2.54 $\pm$ 0.46 \\
nnFormer~\cite{zhou2023nnformer}  & 91.44 $\pm$ 0.95 & 2.12 $\pm$ 1.11 \\
\cmidrule(r){1-3}
MIM~\cite{chen2023masked} & 91.68 $\pm$ 0.82 & 2.34 $\pm$ 0.27 \\
MedMAE~\cite{zhou2023self} & 92.45 $\pm$ 0.62 & 2.06 $\pm$ 0.28 \\
RPC-Loc~\cite{lei2021contrastive} & 90.55 $\pm$ 0.46 & 2.57 $\pm$ 0.32\\
VoCo~\cite{wu2024voco} & 92.65 $\pm$ 0.29 & 2.38 $\pm$ 0.60\\
PCRLv2~\cite{zhou2023unified} & 91.70 $\pm$ 0.57 & 1.87 $\pm$ 0.35 \\
vox2vec~\cite{goncharov2023vox2vec} & \underline{93.84 $\pm$ 0.35} & \underline{1.30 $\pm$ 0.37} \\
\cmidrule(r){1-3}
\rowcolor{myblue}
ours & \textbf{94.58 $\pm$ 0.14} & \textbf{0.58 $\pm$ 0.12} \\
\bottomrule
\end{tabular} 
\end{minipage}

\vspace{3mm}

\begin{minipage}{0.32\textwidth}
\subcaption{OASIS1} 
\centering
\scriptsize
\makeatletter\def\@captype{table}\makeatother
\label{}
\begin{tabular}{lcc}
\toprule
& \textbf{DSC(\%)$\uparrow$} & \textbf{HD95(mm)$\downarrow$} \\
\cmidrule(r){1-3}
UNet~\cite{ronneberger2015u}  & 85.98 $\pm$ 0.67 & 0.77 $\pm$ 0.46 \\
nnUNet~\cite{isensee2021nnu}  & 89.51 $\pm$ 0.33 & 0.73 $\pm$ 0.64 \\
SwinUNETR~\cite{hatamizadeh2021swin}  & 88.36 $\pm$ 0.45 & 1.23 $\pm$ 0.80 \\
nnFormer~\cite{zhou2023nnformer}  & 89.62 $\pm$ 0.51 & 0.92 $\pm$ 0.25 \\
\cmidrule(r){1-3}
MIM~\cite{chen2023masked} & 90.36 $\pm$ 0.32 & 0.89 $\pm$ 0.30 \\
MedMAE~\cite{zhou2023self} & 91.89 $\pm$ 0.21 & 0.76 $\pm$ 0.42 \\
RPC-Loc~\cite{lei2021contrastive} & 86.78 $\pm$ 0.71 & 0.97 $\pm$ 0.53\\
VoCo~\cite{wu2024voco} & 92.57 $\pm$ 0.33 & 0.67 $\pm$ 0.45 \\
PCRLv2~\cite{zhou2023unified} & 92.55 $\pm$ 0.17 & \textbf{0.60 $\pm$ 0.36} \\
vox2vec~\cite{goncharov2023vox2vec} & \underline{92.62 $\pm$ 0.51} & 0.92 $\pm$ 0.25 \\
\cmidrule(r){1-3}
\rowcolor{myblue}
ours & \textbf{94.56 $\pm$ 0.14} & \underline{0.60 $\pm$ 0.48} \\
\bottomrule
\end{tabular} 
\end{minipage}\quad
\begin{minipage}{0.32\textwidth}
\subcaption{MSD-Heart}
\centering
\scriptsize
\makeatletter\def\@captype{table}\makeatother
\label{}
\begin{tabular}{lcc}
\toprule
& \textbf{DSC(\%)$\uparrow$} & \textbf{HD95(mm)$\downarrow$} \\
\cmidrule(r){1-3}
UNet~\cite{ronneberger2015u}  & 87.23 $\pm$ 0.17 & 2.21 $\pm$ 0.76 \\
nnUNet~\cite{isensee2021nnu}  & 87.46 $\pm$ 0.33 & 1.33 $\pm$ 0.98 \\
SwinUNETR~\cite{hatamizadeh2021swin}  & 88.04 $\pm$ 0.12 & 1.33 $\pm$ 0.51 \\
nnFormer~\cite{zhou2023nnformer}  & 88.48 $\pm$ 0.21 & 1.62 $\pm$ 0.48 \\
\cmidrule(r){1-3}
MIM~\cite{chen2023masked} & 87.52 $\pm$ 0.48 & 1.55 $\pm$ 0.29 \\
MedMAE~\cite{zhou2023self} & 88.76 $\pm$ 0.17 & 1.60 $\pm$ 0.37 \\
RPC-Loc~\cite{lei2021contrastive} & 87.13 $\pm$ 0.23 & 2.12 $\pm$ 0.26\\
VoCo~\cite{wu2024voco} & 88.89 $\pm$ 0.31 & 1.07 $\pm$ 0.22\\
PCRLv2~\cite{zhou2023unified} & 88.02 $\pm$ 0.07 & \underline{1.10 $\pm$ 0.54} \\
vox2vec~\cite{goncharov2023vox2vec} & \underline{90.13 $\pm$ 0.23} & 1.14 $\pm$ 0.23 \\
\cmidrule(r){1-3}
\rowcolor{myblue}
ours & \textbf{90.55 $\pm$ 0.35} & \textbf{1.05 $\pm$ 0.17} \\
\bottomrule
\end{tabular} 
\end{minipage}\quad
\begin{minipage}{0.32\textwidth}
\subcaption{ADNI}
\centering
\scriptsize
\makeatletter\def\@captype{table}\makeatother
\label{}
\begin{tabular}{lcc}
\toprule
& \textbf{DSC(\%)$\uparrow$} & \textbf{HD95(mm)$\downarrow$} \\
\cmidrule(r){1-3}
UNet~\cite{ronneberger2015u}  & 80.34 $\pm$ 0.48 & 3.21 $\pm$ 0.98  \\
nnUNet~\cite{isensee2021nnu}  & 84.77 $\pm$ 0.74 & 2.35 $\pm$ 0.80  \\
SwinUNETR~\cite{hatamizadeh2021swin}  & 84.98 $\pm$ 0.93 & 2.12 $\pm$ 0.47  \\
nnFormer~\cite{zhou2023nnformer}  & 85.60 $\pm$ 0.49 & 1.15 $\pm$ 0.30 \\
\cmidrule(r){1-3}
MIM~\cite{chen2023masked} & 86.02 $\pm$ 0.44 & 1.09 $\pm$ 0.19 \\
MedMAE~\cite{zhou2023self} & 86.87 $\pm$ 0.40 & \underline{1.04 $\pm$ 0.15} \\
RPC-Loc~\cite{lei2021contrastive} & 82.13 $\pm$ 0.67 & 2.33 $\pm$ 0.97\\
VoCo~\cite{wu2024voco} & 86.76 $\pm$ 0.54 & 2.64 $\pm$ 0.60\\
PCRLv2~\cite{zhou2023unified} & 87.76 $\pm$ 0.38 & 1.21 $\pm$ 0.48 \\
vox2vec~\cite{goncharov2023vox2vec} & \underline{87.91 $\pm$ 0.40} & 1.12 $\pm$ 0.23 \\
\cmidrule(r){1-3}
\rowcolor{myblue}
ours & \textbf{88.64 $\pm$ 0.37} & \textbf{1.00 $\pm$ 0.15} \\
\bottomrule
\end{tabular} 
\end{minipage}

\vspace{3mm}

\begin{minipage}{0.32\textwidth}
\subcaption{VS} 
\centering
\scriptsize
\makeatletter\def\@captype{table}\makeatother
\label{}
\begin{tabular}{lcc}
\toprule
& \textbf{DSC(\%)$\uparrow$} & \textbf{HD95(mm)$\downarrow$} \\
\cmidrule(r){1-3}
UNet~\cite{ronneberger2015u}  & 79.33 $\pm$ 0.54 & 10.21 $\pm$ 0.97 \\
nnUNet~\cite{isensee2021nnu}  & 81.54 $\pm$ 0.71 & 8.43 $\pm$ 0.65 \\
SwinUNETR~\cite{hatamizadeh2021swin}  & 83.52 $\pm$ 0.42 & 8.34 $\pm$ 0.63 \\
nnFormer~\cite{zhou2023nnformer}  & 83.52 $\pm$ 0.11 & 9.57 $\pm$ 0.77 \\
\cmidrule(r){1-3}
MIM~\cite{chen2023masked} & 83.29 $\pm$ 0.13 & 9.28 $\pm$ 0.70 \\
MedMAE~\cite{zhou2023self} & 84.16 $\pm$ 0.10 & \underline{8.22 $\pm$ 0.48} \\
RPC-Loc~\cite{lei2021contrastive} & 80.64 $\pm$ 0.78 & 9.08 $\pm$ 1.14\\
VoCo~\cite{wu2024voco} & 84.87 $\pm$ 0.34 & 9.94 $\pm$ 0.71\\
PCRLv2~\cite{zhou2023unified} & \underline{85.82 $\pm$ 0.19} & 10.46 $\pm$ 0.94 \\
vox2vec~\cite{goncharov2023vox2vec} & 83.61 $\pm$ 0.09 & 9.93 $\pm$ 0.81 \\
\cmidrule(r){1-3}
\rowcolor{myblue}
ours & \textbf{86.70 $\pm$ 0.13} & \textbf{5.60 $\pm$ 0.28} \\
\bottomrule
\end{tabular} 
\end{minipage}\quad
\begin{minipage}{0.32\textwidth}
\subcaption{AbdomenCT-1K} 
\centering
\scriptsize
\makeatletter\def\@captype{table}\makeatother
\label{}
\begin{tabular}{lcc}
\toprule
& \textbf{DSC(\%)$\uparrow$} & \textbf{HD95(mm)$\downarrow$} \\
\cmidrule(r){1-3}
UNet~\cite{ronneberger2015u}  & 71.53 $\pm$ 0.86 & 3.78 $\pm$ 0.83 \\
nnUNet~\cite{isensee2021nnu}  & 73.94 $\pm$ 0.75 & 2.98 $\pm$ 0.57 \\
SwinUNETR~\cite{hatamizadeh2021swin}  & 73.73 $\pm$ 0.45 & 3.16 $\pm$ 0.93 \\
nnFormer~\cite{zhou2023nnformer}  & 74.23 $\pm$ 0.55 & 3.96 $\pm$ 0.82 \\
\cmidrule(r){1-3}
MIM~\cite{chen2023masked} & 74.78 $\pm$ 0.57 & 4.04 $\pm$ 0.95 \\
MedMAE~\cite{zhou2023self} & 76.52 $\pm$ 0.46 & 3.73 $\pm$ 0.73 \\
RPC-Loc~\cite{lei2021contrastive} & 73.54 $\pm$ 0.36 & 4.73 $\pm$ 0.91\\
VoCo~\cite{wu2024voco} & 78.45 $\pm$ 0.64 & 3.29 $\pm$ 0.48\\
PCRLv2~\cite{zhou2023unified} & 76.32 $\pm$ 0.82 & 3.66 $\pm$ 0.70 \\
vox2vec~\cite{goncharov2023vox2vec} & \underline{79.78 $\pm$ 1.45} & \textbf{2.80 $\pm$ 0.80} \\
\cmidrule(r){1-3}
\rowcolor{myblue}
ours & \textbf{79.83 $\pm$ 1.59} & \underline{3.11 $\pm$ 0.71} \\
\bottomrule
\end{tabular} 
\end{minipage}\quad
\begin{minipage}{0.32\textwidth}
\subcaption{LiTs}
\centering
\scriptsize
\makeatletter\def\@captype{table}\makeatother
\label{}
\begin{tabular}{lcc}
\toprule
& \textbf{DSC(\%)$\uparrow$} & \textbf{HD95(mm)$\downarrow$} \\
\cmidrule(r){1-3}
UNet~\cite{ronneberger2015u}  & 79.33 $\pm$ 0.92 & 4.87 $\pm$ 0.62 \\
nnUNet~\cite{isensee2021nnu}  & 80.57 $\pm$ 0.48 & 4.59 $\pm$ 0.87 \\
SwinUNETR~\cite{hatamizadeh2021swin}  & 80.84 $\pm$ 0.59 & 3.62 $\pm$ 0.71 \\
nnFormer~\cite{zhou2023nnformer}  & 81.84 $\pm$ 0.74 & 5.63 $\pm$ 0.22 \\
\cmidrule(r){1-3}
MIM~\cite{chen2023masked} & 80.73 $\pm$ 0.62 & 5.55 $\pm$ 0.92 \\
MedMAE~\cite{zhou2023self} & 83.18 $\pm$ 0.39 & 4.09 $\pm$ 0.79 \\
RPC-Loc~\cite{lei2021contrastive} & 80.33 $\pm$ 0.91 & 5.06 $\pm$ 0.67\\
VoCo~\cite{wu2024voco} & 83.37 $\pm$ 0.62 & 4.43 $\pm$ 0.73\\
PCRLv2~\cite{zhou2023unified} & 82.37 $\pm$ 0.42 & 4.77 $\pm$ 0.45 \\
vox2vec~\cite{goncharov2023vox2vec} & \underline{86.82 $\pm$ 0.34} & \underline{3.21 $\pm$ 0.48} \\
\cmidrule(r){1-3}
\rowcolor{myblue}
ours & \textbf{87.83 $\pm$ 0.28} & \textbf{3.17 $\pm$ 0.35} \\
\bottomrule
\end{tabular}
\end{minipage}

\vspace{3mm}

\begin{minipage}{0.32\textwidth}
\subcaption{MSD-Liver} 
\centering
\scriptsize
\makeatletter\def\@captype{table}\makeatother
\label{}
\begin{tabular}{lcc}
\toprule
& \textbf{DSC(\%)$\uparrow$} & \textbf{HD95(mm)$\downarrow$} \\
\cmidrule(r){1-3}
UNet~\cite{ronneberger2015u}  & 78.42 $\pm$ 0.68 & 5.21 $\pm$ 0.92 \\
nnUNet~\cite{isensee2021nnu}  & 82.15 $\pm$ 0.51 & 4.33 $\pm$ 0.74 \\
SwinUNETR~\cite{hatamizadeh2021swin}  & 83.07 $\pm$ 0.63 & 4.02 $\pm$ 0.81 \\
nnFormer~\cite{zhou2023nnformer}  & 83.92 $\pm$ 0.47 & 3.87 $\pm$ 0.55 \\
\cmidrule(r){1-3}
MIM~\cite{chen2023masked} & 84.36 $\pm$ 0.39 & 3.62 $\pm$ 0.48 \\
MedMAE~\cite{zhou2023self} & 85.24 $\pm$ 0.52 & 3.35 $\pm$ 0.42 \\
RPC-Loc~\cite{lei2021contrastive} & 82.78 $\pm$ 0.71 & 4.23 $\pm$ 0.67 \\
VoCo~\cite{wu2024voco} & 85.63 $\pm$ 0.44 & 3.28 $\pm$ 0.39 \\
PCRLv2~\cite{zhou2023unified} & 86.02 $\pm$ 0.33 & 3.18 $\pm$ 0.41 \\
vox2vec~\cite{goncharov2023vox2vec} & \underline{87.25 $\pm$ 0.36} & \underline{2.95 $\pm$ 0.32} \\
\cmidrule(r){1-3}
\rowcolor{myblue}
ours & \textbf{88.41 $\pm$ 0.29} & \textbf{2.63 $\pm$ 0.26} \\
\bottomrule
\end{tabular} 
\end{minipage}\quad
\begin{minipage}{0.32\textwidth}
\subcaption{MSD-Colon}
\centering
\scriptsize
\makeatletter\def\@captype{table}\makeatother
\label{}
\begin{tabular}{lcc}
\toprule
& \textbf{DSC(\%)$\uparrow$} & \textbf{HD95(mm)$\downarrow$} \\
\cmidrule(r){1-3}
UNet~\cite{ronneberger2015u}  & 72.18 $\pm$ 1.02 & 8.35 $\pm$ 0.87 \\
nnUNet~\cite{isensee2021nnu}  & 75.26 $\pm$ 0.93 & 7.41 $\pm$ 0.92 \\
SwinUNETR~\cite{hatamizadeh2021swin}  & 76.83 $\pm$ 0.81 & 7.02 $\pm$ 0.74 \\
nnFormer~\cite{zhou2023nnformer}  & 77.55 $\pm$ 0.77 & 6.88 $\pm$ 0.63 \\
\cmidrule(r){1-3}
MIM~\cite{chen2023masked} & 78.21 $\pm$ 0.65 & 6.53 $\pm$ 0.71 \\
MedMAE~\cite{zhou2023self} & 79.34 $\pm$ 0.58 & 6.24 $\pm$ 0.68 \\
RPC-Loc~\cite{lei2021contrastive} & 76.92 $\pm$ 0.84 & 7.15 $\pm$ 0.79 \\
VoCo~\cite{wu2024voco} & 80.16 $\pm$ 0.52 & 5.97 $\pm$ 0.55 \\
PCRLv2~\cite{zhou2023unified} & 80.87 $\pm$ 0.61 & 5.63 $\pm$ 0.49 \\
vox2vec~\cite{goncharov2023vox2vec} & \underline{82.35 $\pm$ 0.47} & \underline{4.82 $\pm$ 0.42} \\
\cmidrule(r){1-3}
\rowcolor{myblue}
ours & \textbf{83.76 $\pm$ 0.38} & \textbf{4.13 $\pm$ 0.36} \\
\bottomrule
\end{tabular} 
\end{minipage}

\end{table*}

\subsubsection{Multi-Organs or Tumors Segmentation}

For the 3D segmentation task, we implement this task that includes cardiac structure segmentation (MM-WHS~\cite{zhuang2016multi} and MSD-Heart~\cite{antonelli2022medical}), abdominal organ segmentation (MSD-Liver~\cite{antonelli2022medical}, CHAOS~\cite{kavur2021chaos}, and AbdomenCT-1K~\cite{ma2021abdomenct}), brain segmentation (OASIS1~\cite{marcus2007open} and ADNI~\cite{mueller2005alzheimer}), and tumor segmentation (MSD-Colon~\cite{antonelli2022medical}, VS~\cite{shapey2021segmentation}, LiTs~\cite{bilic2023liver}). As with previous methods~\cite{goncharov2023vox2vec}, we fine-tune the pre-trained 3D encoder and decoder and jointly train a segmentation head using the Dice loss, and use Dice score (DSC) and 95\% Hausdorff distance (HD95) as evaluation metrics.

The 3DSeg-8 dataset~\cite{chen2019med3d}, used for pretraining, includes five tasks that overlap with the MSD benchmark (Hippocampus, Prostate, Pancreas, Vessel, and Spleen). To eliminate domain familiarity and prevent potential data leakage, we exclude these overlapping tasks during evaluation. Instead, we select the Heart, Liver, and Colon tasks from MSD, which are unseen during pretraining, as the downstream evaluation target. This design ensures that performance reflects true cross-domain generalization rather than memorization effects from the pretraining data.

\subsubsection{Disease Classification}
For Alzheimer's disease classification, we use the OASIS2 dataset~\cite{marcus2010open}, consisting of 312 T1-weighted structural MRI scans collected from 135 subjects, including both AD subjects and healthy volunteers. Each MRI scan's image size is $224\times 224 \times 224$, and the voxel spacing is 1.75 mm. 
For Parkinson's disease classification, the PPMI dataset~\cite{marek2011parkinson} consists of 582 T1-weighted structural MRI scans collected from 247 NCs and 335 PDs. Each MRI scan's image size is $256\times 256\times 196$, and the voxel spacing is 1 mm.
We fine-tune the pre-trained 3D encoder, jointly train a classification head using the cross-entropy loss, and use accuracy and AUC as evaluation metrics.

Since the image volumes used in the downstream datasets have various image resolutions and spacing, all image volumes are re-sampled into the isotropic voxel spacing of 1.0 mm in each dimension, and the image size of volumes ranges from $144\times 144\times 161$ to $400\times 400\times 256$. The voxel intensities of the images are then normalized to the range [0,1]. To simplify the preprocessing step, all images are first padded to a cube shape and then scaled to a unified size of $128\times 128\times 128$ as uniform inputs, differs from pretraining's retaining native resolutions. We {\it subject-wisely} split the data into training, validation, and testing with a ratio of 7:1:2.

\begin{table}[htbp]
\begin{center}
\begin{threeparttable} 
\caption{Comparison of our Med3DInsight and baselines on 3D Classification. 
}
\label{tab:results_cls}
\begin{tabular}{lcc} 
\toprule
\multirow{2}*{Methods}   & \textbf{Accuracy(\%)$\uparrow$} & \textbf{AUC(\%)$\uparrow$} \\
\cmidrule(r){2-3}
& \multicolumn{2}{c}{OASIS2}\\
\cmidrule(r){1-3}
Med3D~\cite{chen2019med3d}  & 80.44 $\pm$ 0.76  & 82.32 $\pm$ 0.86 \\
M3T~\cite{jang2022m3t}  & 81.37 $\pm$ 0.43  & 83.48 $\pm$ 0.75 \\
nnFormer~\cite{zhou2023nnformer} & 82.62 $\pm$ 0.57  & 84.43 $\pm$ 0.37  \\
\cmidrule(r){1-3}
MIM$^*$~\cite{chen2023masked} & 82.86 $\pm$ 0.75 & 85.04 $\pm$ 0.25 \\
MedMAE$^*$~\cite{zhou2023self} & 84.15 $\pm$ 0.36 & 86.68 $\pm$ 0.64 \\ 
RPC-Loc$^*$~\cite{lei2021contrastive} & 81.04 $\pm$ 0.35  & 83.34 $\pm$ 0.56 \\
VoCo$^*$~\cite{wu2024voco} & 83.65 $\pm$ 0.33  & 86.23 $\pm$ 0.26\\
PCRLv2$^*$~\cite{zhou2023unified} & \underline{84.71 $\pm$ 0.06} & \underline{87.92 $\pm$ 0.35} \\
vox2vec$^*$~\cite{goncharov2023vox2vec} & 84.14 $\pm$ 0.34 & 87.57 $\pm$ 0.36 \\
\cmidrule(r){1-3}
\rowcolor{myblue}
ours$^*$ & \textbf{86.93 $\pm$ 0.16} & \textbf{89.52 $\pm$ 0.05} \\ 
\cmidrule(r){1-3}
\morecmidrules
\cmidrule(r){1-3} 
  & \multicolumn{2}{c}{PPMI} \\
\cmidrule(r){1-3}
Med3D~\cite{chen2019med3d}  & 76.45 $\pm$ 0.85 & 77.72 $\pm$ 0.37 \\
M3T~\cite{jang2022m3t} & 76.02 $\pm$ 0.63 & 78.04 $\pm$ 0.67 \\
nnFormer~\cite{zhou2023nnformer} & 80.43 $\pm$ 0.64  & 83.22 $\pm$ 0.45  \\
\cmidrule(r){1-3}
MIM$^*$~\cite{chen2023masked} & 80.53 $\pm$ 0.53 & 81.46 $\pm$ 0.26 \\
MedMAE$^*$~\cite{zhou2023self} & 81.84 $\pm$ 0.55 & 83.46 $\pm$ 0.44 \\ 
RPC-Loc$^*$~\cite{lei2021contrastive} & 79.16 $\pm$ 0.76 & 80.04 $\pm$ 0.89\\
VoCo$^*$~\cite{wu2024voco} & 83.12 $\pm$ 0.24 & 85.04 $\pm$ 0.25\\
PCRLv2$^*$~\cite{zhou2023unified} & 82.86 $\pm$ 0.57 & 84.74 $\pm$ 0.22 \\
vox2vec$^*$~\cite{goncharov2023vox2vec} & \underline{83.43 $\pm$ 0.32} & \underline{85.12 $\pm$ 0.27} \\
\cmidrule(r){1-3}
\rowcolor{myblue}
ours$^*$ & \textbf{84.25 $\pm$ 0.32} & \textbf{86.63 $\pm$ 0.15} \\ 
\bottomrule
\end{tabular}
\begin{tablenotes}
\small
\item[*] Use the pre-trained encoder of the model for classification.
\end{tablenotes}
\end{threeparttable}
\end{center}
\end{table}

\subsection{Baselines and Experimental Results}

We conduct extensive experiments on downstream applications involving organ-wise and substructure-wise segmentation tasks as well as classification tasks. During the pretraining phase, we train the 3D visual encoder-decoder using our triplet dataset.
In the fine-tuning phase, we concurrently update the parameters of the pre-trained vision encoder-decoder. 
Our framework is compared with four 3D supervised baselines, i.e., UNet~\cite{ronneberger2015u}, nnUNet~\cite{isensee2021nnu}, SwinUNETR~\cite{hatamizadeh2021swin}, and nnFormer~\cite{zhou2023nnformer}, and six SOTA SSL methods, including two reconstruction-based methods, i.e., MIM~\cite{chen2023masked} and medical MAE~\cite{zhou2023self}, and four alignment-based methods, i.e., RPC-Loc~\cite{lei2021contrastive}, VoCo~\cite{wu2024voco}, PCRLv2~\cite{zhou2023unified} and vox2vec~\cite{goncharov2023vox2vec}.
All SSL methods employed identical fine-tuning procedures, including pretraining hyperparameters (learning rate, batch size, training epochs) pretrain datasets, and evaluation metrics, to eliminate confounding factors from training strategy variations.

\subsubsection{Experimental Results for 3D Segmentation}
\begin{figure}[t]
\begin{center}
\includegraphics[width=\linewidth]{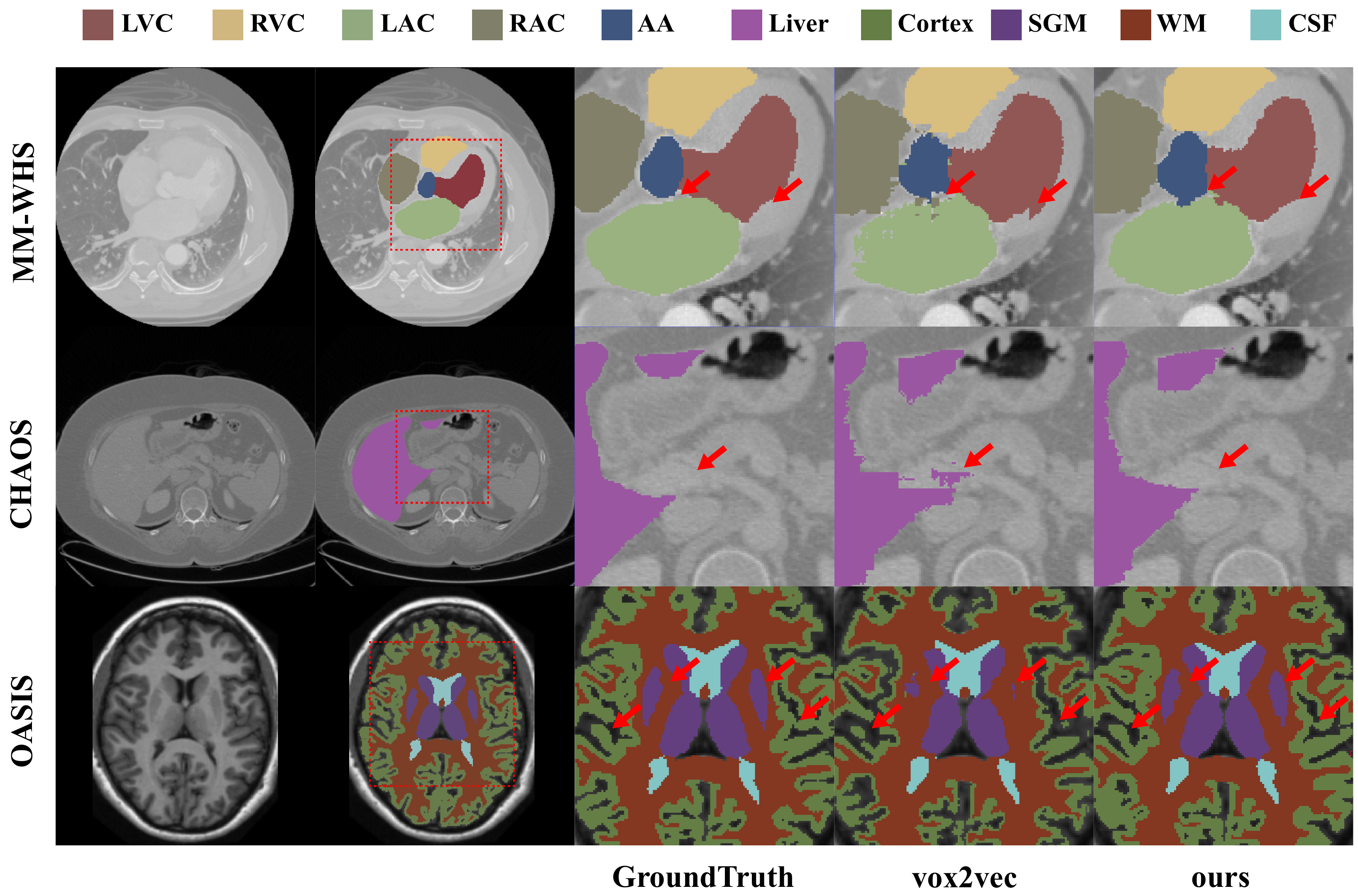}
\caption{Qualitative comparison between our Med3DInsight and vox2vec~\cite{goncharov2023vox2vec} on 3D segmentation tasks. 
}
\label{fig:vis}
\end{center}
\end{figure}

Table~\ref{tab:seg} showcases the experimental results of multi-organ and substructure segmentation across the various datasets, which include 19 different organs or tumors. Our Med3DInsight consistently outperforms SOTA SSL methods over eight datasets, surpassing the best baseline by over 1\% in average Dice score and 0.7mm in average HD95. Our method outperforms other baselines even in scenarios with limited training data in MM-WHS, underscoring the robustness and adaptability of our approach. Also, Med3DInsight demonstrates impressive transferability despite the switch from CT to MRI and from abdominal and cardiac imaging to brain imaging, highlighting its adaptability across varied modalities and anatomical structures. By incorporating clinical knowledge during the pre-training stage, Med3DInsight learns 3D visual features that significantly enhance its performance in unseen domains and with different organs. 
Fig.~\ref{fig:vis} visualizes the qualitative comparison of 3D segmentation results compared with the SOTA SSL baseline vox2vec~\cite{goncharov2023vox2vec}. Med3DInsight  generates segmentation results more consistent with the ground truth, particularly in capturing fine-grained details of organs or substructures, and exhibits higher integrity in segmentation.

\subsubsection{Experimental Results for 3D Classification}
Table~\ref{tab:results_cls} reports the 3D classification results on the OASIS2 dataset and PPMI dataset. The OASIS2 dataset presents the challenge of discerning Alzheimer's disease, and the PPMI dataset presents the challenge of discerning Parkinson's disease. Nevertheless, our approach consistently excels in recognizing pathological conditions, achieving the highest accuracy and AUC scores across all baselines and other SSL methods. This superiority can be attributed to the inherent advantages of the semantic comprehension ability of large language models. Unlike SSL, which primarily learns visual invariance from augmented views, VLP leverages the rich clinical knowledge encoded in text, providing a more informed supervision signal. This experiment demonstrates that VLP, guided by MLLMs with medical image explanation knowledge, offers more semantically representative 3D visual features than vision-only SSL. 

\noindent \textbf{Med3DInsight vs. 3D/Pretrained Models.} 
Unlike conventional 3D or 3D-pretrained models that focus solely on anatomical structures, Med3DInsight leverages image-text pre-training to embed clinical semantics into the visual representation space for both image segmentation and classification tasks. This semantic alignment enables the model to capture structure- or disease-relevant features that go beyond image intensity, providing contextual cues, especially beneficial when annotations are limited or ambiguous. Table~\ref{tab:seg} and~\ref{tab:results_cls} consistently report that Med3DInsight outperforms purely visual 3D or 3D pretrained models, demonstrating the advantage of integrating clinical language priors. Ultimately, Med3DInsight establishes a new paradigm for medical vision pre-training by unifying three critical dimensions: (1) anatomical structural awareness through 3D context modeling, (2) disease semantic understanding via language-guided supervision, and (3) cross-modal adaptability enabled by modality-agnostic representation learning. This tripartite synergy explains its consistent superiority across segmentation and classification tasks.

\subsection{Ablation Study and Other Analysis}

\subsubsection{Effectiveness of the PSAT Model}
To demonstrate the effectiveness of each component in PSAT, we conduct an ablation study on the OASIS1 3D segmentation task. In Table~\ref{tab:ablation}(a), $Ex1$ represents the model without any query transformer or place-slice position embedding. In this scenario, the 3D image feature is directly fed into a projection layer to align the dimensions, and then the contrastive loss is calculated to align with the 2D vision-language features of CLIP.
Using the query transformer, we observe an average dice score of 93.93\%, indicating that the query transformer better aligns 3D medical image features with 2D vision-language features.
The introduction of the Plane-Slice Position embedding results in a 0.82\% average improvement in the dice score. This result confirms that incorporating the Plane-Slice Position embedding enhances the model's understanding of 3D scenes and spatial relationships.

\begin{figure*}[t]
        \centering
		\includegraphics[width=\linewidth]{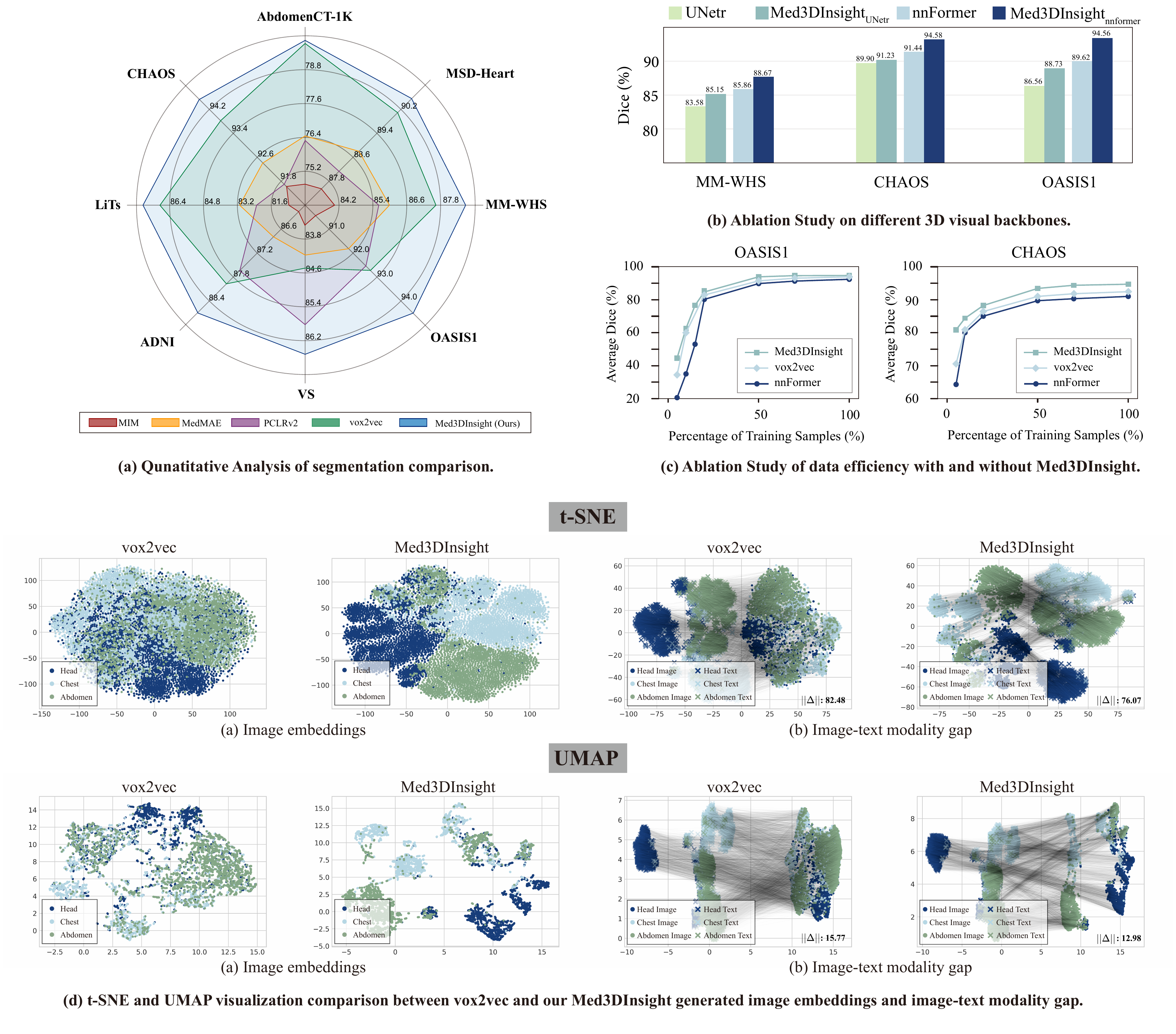}
		\caption{Qunatitative and Qualitative Analysis of our method with existing visual SSL methods. 
  (a) Segmentation comparison of our method with existing four visual SSL methods across eight different 3D medical image segmentation datasets. Results are reported in terms of Dice scores (\%).
  (b) Ablation study of different 3D visual backbones.
  (c) Ablation study of data efficiency with and without Med3DInsight.
  (d) t-SNE and UMAP visualization comparison between vox2vec and our Med3DInsight generated image embeddings and image-text modality gap. The black lines refer to image-text pairs.
  }
		\label{fig:overall}
\end{figure*}

\subsubsection{Loss Ablation}
We further investigate the influence of different losses during the pre-training stage. As depicted in Table~\ref{tab:ablation}(b), excluding either $\mathcal{L}_{\text{mPOT}}$ or $\mathcal{L}_{\text{res}}$ leads to reduced performance. However, when combining all components as in our method, we achieve the highest performance. This underscores the importance of both losses for enhancing the effectiveness of the proposed method. 

\subsubsection{Alignment Strategy}
We conduct experiments on the OASIS1 test set to examine the transferability of the learned joint embedding space. We report dice scores for contrastive loss and the mini-batch partial optimal transport loss in Table~\ref{tab:ablation}(c). The mPOT loss acquires the highest dice score for the segmentation task, which means the proposed mPOT loss achieves the smallest gap between 3D and 2D embedding, which means it is more transferable to downstream tasks than contrastive loss.

\subsubsection{Vision/Text Encoder}
To assess the versatility of our method, we employ two distinct 2D encoders: the CLIP~\cite{radford2021learning} and the BioMedCLIP~\cite{zhang2023biomedclip}. As shown in Table~\ref{tab:ablation}(d), the performance differences between these encoders are marginal, underscoring that our method is agnostic to these two choices of backbones for vision and text encoders.

\begin{table}[t]
\begin{center}
    \caption{Ablation Experiments, reporting Dice Scores (\%) on OASIS1 dataset. QTrans denotes the Query Transformer, and PSPE denotes the Plane-Slice Position Embedding.
    }
    \label{tab:ablation}
    \begin{tabular}{lccccc}
    \cmidrule(r){1-6}
        Task & Cortex & SGM & WM & CSF & Avg. \\
    \cmidrule(r){1-6}
        \textbf{(a) PSAT} \\
    \cmidrule(r){1-6}
        w/o QTrans & 88.54 & 92.79 & 94.60 & 93.92 & 92.46 \\
        QTrans & 90.76 & 94.14 & 95.66 & 95.19 & 93.93  \\
        QTrans + PSPE & \textbf{92.07} & \textbf{95.23} & \textbf{96.13} & \textbf{95.60} & \textbf{94.75} \\
    \cmidrule(r){1-6}
        \textbf{(b) Loss Components} \\
    \cmidrule(r){1-6}
        $\mathcal{L}_{\text{res}}$ & 87.51 & 91.97 & 94.20 & 93.50 & 91.79 \\
        $\mathcal{L}_{\text{mPOT}}$ & 90.65 & 94.01 & 95.56 & 95.15 & 93.84 \\
        $\mathcal{L}_{\text{res}}$ + $\mathcal{L}_{\text{mPOT}}$ & \textbf{92.07} & \textbf{95.23} & \textbf{96.13} & \textbf{95.60} & \textbf{94.75} \\
    \cmidrule(r){1-6}
        \textbf{(c) Alignement Strategy} \\
    \cmidrule(r){1-6}
        Contrastive loss & 90.63 & 94.08 & 95.56 & 95.22 & 93.87 \\
        mPOT loss & \textbf{92.07} & \textbf{95.23} & \textbf{96.13} & \textbf{95.60} & \textbf{94.75} \\
    \cmidrule(r){1-6}
        \textbf{(d) Vision/Text Encoder} \\
    \cmidrule(r){1-6}
        BioMedCLIP & 92.01 & \textbf{95.26} & 96.11 & 95.56 & 94.73 \\ 
        CLIP & \textbf{92.07} & 95.23 & \textbf{96.13} & \textbf{95.60} & \textbf{94.75} \\
    \cmidrule(r){1-6}
        \textbf{(e) 2D MLLMs} \\
    \cmidrule(r){1-6}
        HuatuoGPT-Vision & 90.13 & 93.74 & 95.37 & 94.86 & 93.52 \\ 
        GPT-4Vision & \textbf{92.07} & \textbf{95.23} & \textbf{96.13} & \textbf{95.60} & \textbf{94.75} \\
    \cmidrule(r){1-6}
        \textbf{(f) Slice Per Volume} \\
    \cmidrule(r){1-6}
        5 & 89.54 & 91.79 & 93.60 & 92.92 & 91.96 \\
        3 & 91.76 & 95.14 & 94.67 & \textbf{96.19} & 94.44  \\
        1 & \textbf{92.07} & \textbf{95.23} & \textbf{96.13} & 95.60 & \textbf{94.75} \\
    \cmidrule(r){1-6}
        \textbf{(g) Preprocess Type} \\
    \cmidrule(r){1-6}
        Resize & 92.07 & 95.23 & 96.13 & \textbf{95.60} & 94.75 \\
        Sliding Window & \textbf{93.48} & \textbf{95.47} & \textbf{96.78} & 95.22 & \textbf{95.23} \\
    \cmidrule(r){1-6}
    \end{tabular}
  \end{center}
\end{table}

\begin{table*}[pbt]
\scriptsize
\centering
\caption{Slice description generated by different multimodal LLMs (\textcolor{blue}{Blue}: accurate description; \textcolor{red}{red}: inaccurate).}
\label{tab:mllm}
\begin{tabular}{p{90pt}p{200pt}p{200pt}}
\cmidrule{1-3}
Samples & (a) \begin{minipage}[b]{0.2\columnwidth} \raisebox{-.5\height}{\includegraphics[width=\linewidth]{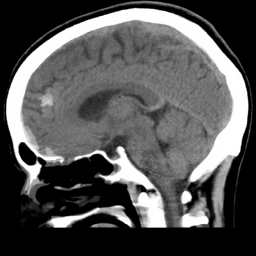}} \end{minipage} & (b) \begin{minipage}[b]{0.2\columnwidth} \raisebox{-.5\height}{\includegraphics[width=\linewidth]{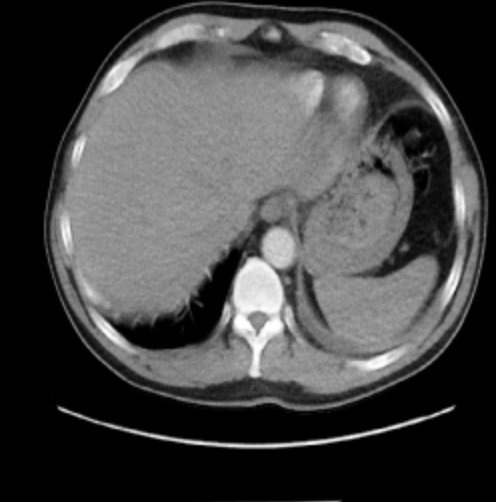}} \end{minipage} \\
\cmidrule{1-3}
    GPT-4Vision~\cite{2023GPT4VisionSC} & This image is a \cBlue{\textbf{sagittal CT scan of the human brain}}, providing a detailed view of the cranial structures. \cBlue{\textbf{The cerebral cortex with its characteristic folds is clearly visible, along with the midline structures such as the corpus callosum and brainstem. The ventricles, which are fluid-filled spaces, appear as darker areas. The bright white outline represents the skull, which encases the brain.}} This type of scan is crucial for assessing brain abnormalities, injuries, or diseases, as it allows for a comprehensive evaluation of the brain's anatomy and potential pathologies. & This CT scan image shows \cBlue{\textbf{an axial view of the abdomen}}, highlighting various internal structures. \cBlue{\textbf{The large organ on the left side is the liver, appearing in a homogeneous gray tone. Adjacent to it, on the right, is the spleen, which is smaller and similarly shaded. The vertebral column is visible at the bottom center, appearing as a white circular structure. Surrounding tissues and organs, including portions of the intestines, are also visible}}, providing a detailed view of the abdominal cavity. This type of scan is essential for diagnosing abdominal conditions and assessing organ health. \\
\cmidrule{1-3}
Claude 3~\cite{TheC3} & The image appears to be a medical scan, likely a CT or MRI, showing a cross-sectional view of a \cBlue{\textbf{human brain. The brain structures are visible in shades of gray, with different densities represented.}} The scan appears to be focused on a specific region of the brain, revealing intricate details of the anatomy. Medical professionals would use this type of image to analyze the brain's structure and identify any potential abnormalities or pathologies that may require further examination or treatment. & This medical image appears to be a cross-sectional computed tomography (CT) \cBlue{\textbf{scan of the abdomen}}. The image shows a detailed view of the internal structures, including the \cBlue{\textbf{liver, spleen, and other abdominal organs}}. The high-contrast, black and white image provides a comprehensive representation of the anatomical features, allowing for the identification and examination of various organs and tissues. This type of imaging is commonly used in medical diagnosis and treatment planning for conditions affecting the abdominal region. \\
\cmidrule{1-3}
HuatuoGPT-Vision-7B~\cite{chen2024huatuogptvisioninjectingmedicalvisual} & The image shows a \cBlue{\textbf{sagittal view of the brain}}, which appears to be normal in structure and anatomy. \cBlue{\textbf{The cerebral cortex, ventricles, cerebellum, and other key structures}} are clearly visible without any obvious abnormalities or pathological findings. This suggests that this is likely an unremarkable CT scan of the head taken after some form of treatment or intervention, as indicated by the reference information provided. & The image shows a computed tomography (CT) \cBlue{\textbf{scan of the abdomen}}, which reveals an \cRed{\textbf{enlarged liver with multiple hypodense lesions throughout both lobes. These findings are consistent with metastatic disease involving the liver parenchyma and peritoneal surfaces.}} The CT scan provides detailed anatomical information that can be used to assess the extent and distribution of the hepatic lesions for further clinical evaluation and management. \\
\cmidrule{1-3}

\end{tabular}
\end{table*}

\subsubsection{Ablation of different image backbones}

As seen in Figure~\ref{fig:overall}(b), we employ two distinct image backbones UNETR~\cite{hatamizadeh2022unetr} and nnFormer~\cite{zhou2023nnformer}. Med3DInsight consistently improves different backbones across three datasets, indicating it is a comprehensive representation learning framework that enriches visual representation for medical image segmentation.

\subsubsection{Ablation of different multimodal large language models}
We conduct a qualitative comparison of the generation quality across three MLLMs, as shown in Table~\ref{tab:mllm}, including GPT-4Vision and Claude, which are general-purpose models, and HuatuoGPT-Vision, which is specifically designed for the medical domain.  in which the GPT-4V we use have more accurate descriptions.
Additionally, we perform a quantitative evaluation of the two state-of-the-art MLLMs, i.e., GPT-4Vision for general tasks and HuatuoGPT-Vision for medical tasks, on the downstream segmentation task on the OASIS1 dataset. As presented in Table~\ref{tab:ablation}(e), the results demonstrate that the quality of the generated descriptions during our pretraining phase closely correlates with the performance of this downstream task.

\subsubsection{Different Slice-per-volume Strategies}
To validate our design choice, we conduct ablation experiments comparing different slice-per-volume sampling strategies during pretraining. Specifically, we test a setting using only the M3D subset (8,000 volumes) with three slices per volume, resulting in the same total number of slices (24,000) as our original approach. Despite this matched slice count, the model trained with higher slice density on fewer volumes underperformed relative to our one-slice-per-volume strategy applied to the combined 3DSeg-8 and M3D datasets (24,000 volumes). As reported in Table~\ref{tab:ablation}(f), additional experiments using 1, 3, and 5 slices per volume showed that adding more slices from the same volume does not provide performance gains, highlighting the redundancy of adjacent slices within the same volume. These findings support two key conclusions: (1) Cross-volume anatomical diversity is more beneficial for generalization than dense intra-volume sampling; and (2) Sampling one representative slice per volume offers an effective trade-off between dataset diversity and computational efficiency.

\subsubsection{Different Preprocess Types}
To evaluate the impact of different image preprocessing strategies, we conduct additional experiments comparing two approaches:
(1) Isotropic resampling (i.e., resizing) of the entire volume to a fixed $128^3$ resolution, as we did;
(2) Sliding-window-based ROI cropping using MONAI~\cite{cardoso2022monai}, which extracts $128^3$ patches from the original $256^3$ volumes.
As shown in Table~\ref{tab:ablation}(g), 
Sliding window achieves slightly higher accuracy for targets with rich edge details, but 
resize provided faster inference with comparable Dice scores. This suggests that uniform resampling may better preserve global anatomical context, offering a modest advantage for segmentation tasks.

\subsubsection{Data Efficiency}
Model pretraining can potentially reduce the demand for labeled data in downstream tasks. To validate the data efficiency of our Med3DInsight framework, we check the average dice changes with varying amounts of fine-tuning samples. Our method consistently outperforms baseline methods across all training data percentages. As shown in Figure~\ref{fig:overall}(c), pretraining under the Med3DInsight framework substantially enhances nnFormer's performance in low-data regimes. Notably, Med3DInsight exhibits superior performance on the CHAOS dataset compared to the OASIS1 dataset when utilizing less than 20\% of the training data. This disparity likely arises from our pretraining datasets, which are rich in abdominal CT images. Nevertheless, Med3DInsight still delivers significant performance improvements on the OASIS1 dataset.

\subsubsection{Latent Representation of Pretrained Model}
We utilize the pretrained visual encoder to extract latent representations from 3D medical volume images of the brain, abdomen, and chest. We then apply the t-SNE algorithm~\cite{van2008visualizing} and UMAP~\cite{becht2019dimensionality} for dimensionality reduction and visualization. Figure~\ref{fig:overall}(d) shows our resulting 2D representations, comparing the SOTA SSL baseline vox2vec~\cite{goncharov2023vox2vec}. The t-SNE visualization shows that in scenarios involving multiple anatomies, our model demonstrates superior class separability compared to vox2vec. This indicates that our MLLM-guided approach effectively captures distinguishing semantic features across human anatomies.

To demonstrate the semantic understanding ability of our 3D visual encoder, we randomly sample 1000 pairs of image volumes and texts from the M3D datasets~\cite{bai2024m3d}.
We extract image embeddings from our encoder and vox2vec, and visualize the modality gap $||\Delta ||$~\cite{liang2022mind} between the embeddings extracted by two image encoders and the text embeddings extracted by BERT~\cite{devlin2018bert}.
Compared to vox2vec, the modality gap of Med3DInsight decreases from 82.48 to 76.07 on t-SNE and from 15.77 to 12.98 on UMAP, as shown in Fig.~\ref{fig:overall}(d), confirming that our model achieves a better visual representation that is closer to the text description.

\section{Discussion}

\subsection{
Does the algorithm extract a single slice or multiple contiguous slices from the 3D image volume?
}
Generating detailed textual descriptions for every 2D slice of a 3D image volume is computationally expensive and unnecessary for a comprehensive understanding of the volume. Our experiments show that as the interval between slices decreases, the resulting descriptions become increasingly redundant. Also, when the slice selection is random, any single slice can be used effectively for extraction and description. Given that the pre-training dataset includes thousands of scans from diverse anatomical regions, e.g., head, chest, abdomen, extracting and describing one slice per 3D volume is sufficient, especially in the context of large-scale datasets.

\subsection{Can medical MLLMs replace GPT-4V in the proposed approach? What are the limitations of GPT-4V?}
Since GPT-4V is a general-purpose multimodal model, we attempted to evaluate whether medical MLLMs could enhance 3D medical image understanding. Our analysis shows that GPT-4V exhibits strong medical comprehension capabilities, even without fine-tuning on specialized medical datasets. Notably, GPT-4V outperforms other available multimodal LLMs, such as Claude 3, in accurately describing detailed anatomical structures. For example, as shown in Table~\ref{tab:mllm}, GPT-4V correctly identifies the bright white region as the skull. In contrast, HuatuoGPT-Vision, despite being fine-tuned on medical data, performs worse than GPT-4V. For instance, in the sample (b) of Table~\ref{tab:mllm}, depicting an abdominal scan with normal upper abdominal organs, HuatuoGPT-Vision incorrectly reports ``multiple hypodense lesions". This discrepancy may stem from its training on PubMed articles, which often include images from studies focused on rare or unusual pathologies, potentially biasing the model toward identifying uncommon conditions not typically seen in routine clinical practice.

While GPT-4V demonstrates certain strengths, it is not specifically designed for complex medical image interpretation or precise medical analysis. For instance, tasks like estimating the volume or size of lesions require accurate spatial perception and the ability to assess 3D attributes, which generally surpass GPT-4V’s capabilities. Fortunately, in most cases, GPT-4V performs well when describing 2D slices, making it suitable for our task. However, the application of foundation models in medical image analysis presents other challenges. There is a potential risk of incorrect interpretations due to biases or errors in the underlying algorithms and the datasets used for model training. For example, GPT-4V has occasionally misidentified a typical heart scan as an abdominal scan, leading to inaccurate conclusions. Another concern is the potential misuse of medical data, which could inadvertently disclose sensitive health information or personal data embedded in the training set. This underscores the need for transparency in understanding how these models function and make decisions in medical contexts, and the need for safeguards to ensure data privacy, accuracy, and security.

\subsection{Training Data and Computational Cost}
Developing large-scale medical datasets with image-text pairs is particularly challenging due to the complexity of generating detailed and context-rich descriptions for 3D medical images. The textual descriptions of 3D medical images are typically sourced from diagnostic reports or medical image databases, which often focus solely on pathology, lacking detailed information about the broader content within the images. As a result, the quality of the text-image pairs for 3D datasets is often inadequate for training robust models. Med3DInsight addresses this issue by employing generative descriptions as part of a self-supervised learning approach, leveraging clinical insights from advanced 2D MLLMs to compensate for the lack of large-scale 3D datasets. Moreover, by analyzing 3D images through slice-based representations, Med3DInsight aligns with the diagnostic practices of clinicians. The slice-level textual descriptions tend to be more detailed and accurate than those derived directly from reports describing the entire 3D volume.

Moreover, training or fine-tuning large models for 3D medical image understanding demands substantial time and computational resources. For instance, M3D~\cite{bai2024m3d} requires pre-training on a vision encoder using eight NVIDIA A100 GPUs (80 GB each) to process a dataset of 120k image-text pairs in parallel. In contrast, Med3DInsight offers a key advantage by eliminating the need for extensive data preparation and significantly reducing both training time and memory costs.

\section{Conclusion}
In this paper, we introduce Med3DInsight, a novel pretraining framework that leverages 2D MLLMs to enhance medical image understanding and improve downstream segmentation and classification performance. To address the feature space disparity between the 3D image encoder and 2D MLLMs, we propose the Plane-Slice-Aware Transformer (PSAT) module, which employs a learnable query mechanism to bridge the gap between different feature dimensions. Furthermore, we propose a modified mini-batch partial optimal transport to tackle the partial alignment issues among 3D image volumes, 2D image slices, and associated inaccurate text descriptions. Our experimental results consistently demonstrate the superiority of our approach in downstream tasks over existing SOTA methods across multiple datasets. Future work will focus on refining the fine-grained semantic understanding of 3D medical images and exploring the integration with large language models to further reduce noises in generated content.




\bibliographystyle{IEEEtran}
\bibliography{refs}

\end{document}